%% file: main.tex
\documentclass[lettersize,journal]{IEEEtran}
\usepackage{amsmath,amsfonts}
\usepackage{algorithmic}
\usepackage{algorithm}
\usepackage{array}
\usepackage[caption=false,font=normalsize,labelfont=sf,textfont=sf]{subfig}
\usepackage{textcomp}
\usepackage{stfloats}
\usepackage{url}
\usepackage{verbatim}
\usepackage{graphicx}
\usepackage{cite}

\usepackage{booktabs}
\usepackage{xcolor}
\definecolor{iccvblue}{rgb}{0.21,0.49,0.74}
\usepackage[pagebackref,breaklinks,colorlinks,allcolors=iccvblue]{hyperref}
\usepackage{multirow}
\usepackage{multicol}
\usepackage{makecell}
\usepackage{pifont} 
\usepackage{bbding}
\usepackage{colortbl}
\usepackage{subcaption}
\definecolor{deepgreen}{rgb}{0.15, 0.65, 0.42} % deep green
\definecolor{selfdomain}{rgb}{0.9, 1, 0.9} % light green
\definecolor{crossdomain}{rgb}{1, 0.95, 0.8} % light yellow
\definecolor{increase}{rgb}{0.773, 0.384, 0.364} % yellow
\definecolor{decrease}{rgb}{0.243, 0.451, 0.820} % blue

\newcommand{\eg}{\emph{e.g.}}
\newcommand{\etal}{\emph{et al.}}

\begin{document}

\title{Improving the generalization of gait recognition with limited datasets}

\author{Qian Zhou\textsuperscript{*},
        Xianda Guo\textsuperscript{*,$\dagger$},
        Jilong Wang,
        Chuanfu Shen,
        Zhongyuan Wang\textsuperscript{$\ddagger$},~\IEEEmembership{Member,~IEEE},
        Zhen Han,
        Qin Zou,~\IEEEmembership{Senior Member,~IEEE},
        and Shiqi Yu,~\IEEEmembership{Member,~IEEE}
        % <-this % stops a space
\IEEEcompsocitemizethanks{Qian Zhou, Xianda Guo, Zhongyuan Wang, Zhen Han, and Qin Zou are with the School of Computer Science, Wuhan University, Wuhan 430072, China (e-mail: zhouqian@whu.edu.cn; xianda\_guo@163.com; wzy\_hope@163.com; hanzhen\_1980@163.com; qinz@whu.edu.cn). }% <-this % stops a space
\IEEEcompsocitemizethanks{Jilong Wang is with the Department of Automation, School of Information Science and Technology, University of Science and Technology of China, Hefei 230026, China.}% (e-mail: jilongw@mail.ustc.edu.cn)
\IEEEcompsocitemizethanks{Chuanfu Shen is with the Shenzhen Institute of Advanced Study, University of Electronic Science and Technology of China (UESTC), Chengdu 610054, China.}%(e-mail: chuanfu.shen@uestc.edu.cn)
\IEEEcompsocitemizethanks{Shiqi Yu is with the Department of Computer Science and Engineering, Southern University of Science and Technology (SUSTech), Shenzhen 518055, China.}%(e-mail: yusq@sustech.edu.cn)
\IEEEcompsocitemizethanks{$^*$ indicates equal contributions. $^\dagger$: Project Leader. $^\ddagger$: Corresponding author.}
}

\markboth{Journal of \LaTeX\ Class Files,~Vol.~14, No.~8, August~2021}%
{Shell \MakeLowercase{\textit{et al.}}: A Sample Article Using IEEEtran.cls for IEEE Journals}


\maketitle

\input{sec/0_abstract}    
\input{sec/1_intro}
\input{sec/2_related_works}

\input{sec/3_method}
\input{sec/4_experiments}

\input{sec/5_conclusion}

\section*{Acknowledgments}
This work was supported by National Natural Science Foundation of China (62371350, 62372339, 62171324), Key Science and Technology Research Project of Xinjiang Production and Construction Corps (2025AB029).

\bibliographystyle{IEEEtran}
\bibliography{IEEEabrv, ref}

\vfill

\end{document}

%% file: sec/0_abstract.tex
\begin{abstract}
Generalized gait recognition remains challenging due to significant domain shifts in viewpoints, appearances, and environments. Mixed-dataset training has recently become a practical route to improve cross-domain robustness, but it introduces underexplored issues: 1) inter-dataset supervision conflicts, which distract identity learning, and 2) redundant or noisy samples, which reduce data efficiency and may reinforce dataset-specific patterns.
To address these challenges, we introduce a unified paradigm for cross-dataset gait learning that simultaneously improves motion-signal quality and supervision consistency. We first increase the reliability of training data by suppressing sequences dominated by redundant gait cycles or unstable silhouettes, guided by representation redundancy and prediction uncertainty. This refinement concentrates learning on informative gait dynamics when mixing heterogeneous datasets. In parallel, we stabilize supervision by disentangling metric learning across datasets, forming triplets within each source to prevent destructive cross-domain gradients while preserving transferable identity cues. These components act in synergy to stabilize optimization and strengthen generalization without modifying network architectures or requiring extra annotations. 
Experiments on CASIA-B, OU-MVLP, Gait3D, and GREW with both GaitBase and DeepGaitV2 backbones consistently show improved cross-domain performance without sacrificing in-domain accuracy. These results demonstrate that data selection and aligning supervision effectively enables scalable mixed-dataset gait learning.
\end{abstract}

%% file: sec/1_intro.tex
\section{Introduction}
\label{sec:intro}

\begin{figure}[t] 
    \centering 
    \includegraphics[width=0.98\linewidth]{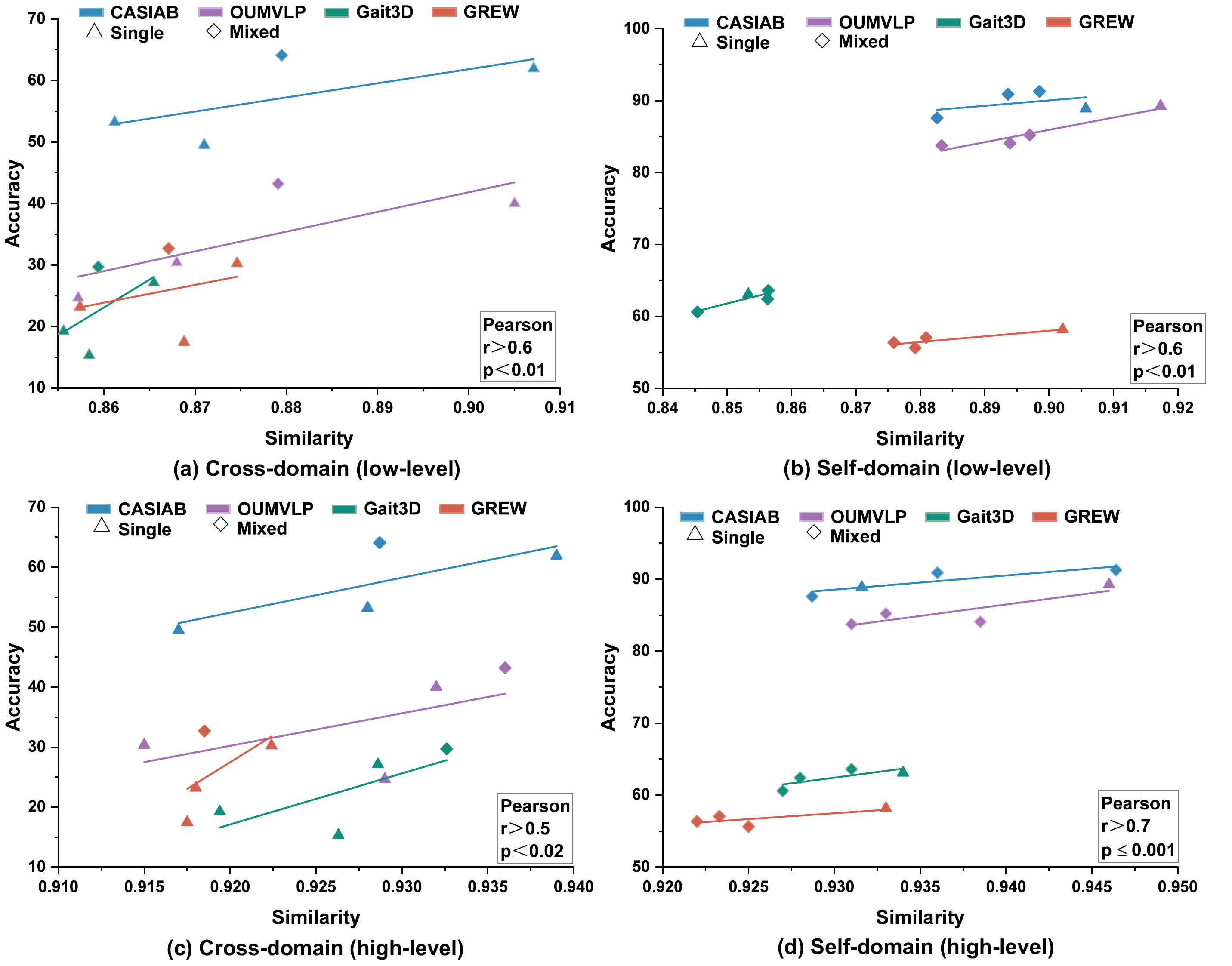} 
    \caption{Relationship between dataset similarity and accuracy across different settings (low/high-level, cross/self-domain) of the GaitBase~\cite{fan2023opengait} model. Mixed training consistently improves performance, especially in cross-domain scenarios. Low-level and high-level indicate pixel-wise and feature-wise similarities, respectively. 
    }
    \label{fig:similarity}
\end{figure}

Gait recognition has gained increasing attention due to its ability to identify individuals from a distance in a contactless manner~\cite{shen2022comprehensive}. Compared to traditional biometrics such as face, iris, and fingerprints, gait enables long-range, non-cooperative identification, making it valuable for security and public safety applications~\cite{filipi2022gait}. With the rapid advancement of deep learning, state-of-the-art gait recognition methods have achieved impressive performance, surpassing 90\%~\cite{fan2023opengait} accuracy on the OU-MVLP dataset~\cite{OUMVLP} and 80\%~\cite{ma2024prompt} on the GREW dataset~\cite{GREW,guo2025gait}, both of which contain thousands of subjects.
\IEEEpubidadjcol

Despite strong in-domain performance, gait models still struggle when deployed across unseen environments, making cross-domain generalization a major bottleneck. Domain shift is a long-standing issue in visual recognition tasks such as person re-identification~\cite{sun2024dualistic}, face analysis~\cite{deng2019arcface}, and action understanding~\cite{lin2024human}, motivating extensive research in domain generalization~\cite{zhou2022domain, wang2022generalizing}. However, gait exhibits amplified sensitivity to domain changes due to its reliance on fine-grained temporal motion patterns and silhouette evolution, which can be easily distorted by variations in viewpoint, clothing, carrying status, walking trajectory, occlusion, and segmentation quality~\cite{sepas2022deep}.

In addition, gait datasets exhibit exceptionally high cross-domain heterogeneity. Controlled indoor datasets such as CASIA-B~\cite{CASIA-B} and OU-MVLP~\cite{OUMVLP} offer clean silhouettes and consistent viewpoints but limited appearance and environmental diversity. In contrast, in-the-wild datasets such as Gait3D~\cite{Gait3D} and GREW~\cite{GREW,guo2025gait} present diverse camera setups, complex motions, cluttered backgrounds, varied resolutions, and substantial segmentation noise. This discrepancy is more pronounced than in face or person re-identification benchmarks, where data collection pipelines are relatively standardized across datasets. As a result, gait models trained on a single dataset tend to memorize dataset-specific motion patterns and scene statistics, rather than learning domain-invariant gait representations, leading to sharp performance degradation under cross-domain evaluation~\cite{wang2025gaitc}.

One practical and straightforward solution to enhance generalization is mixed-dataset training, where multiple datasets are aggregated to expose models to broader data and variations~\cite{ranftl2020towards, shi2024plain}. While effective to a certain extent, naive aggregation introduces two challenges: (1) inter-dataset supervision conflicts that hinder stable optimization, and (2) redundant or noisy samples that degrade representation quality and scalability.

To better understand this phenomenon, we analyze dataset affinity at both appearance and semantic levels. Specifically, low-level similarity is computed from gait energy images (GEIs), while high-level similarity is measured using CLIP embeddings~\cite{radford2021learning}. As shown in Fig.~\ref{fig:similarity}, datasets with higher affinity consistently yield better cross-domain transfer, whereas isolated training on a single dataset results in sharp degradation on dissimilar domains. Mixed training alleviates this effect by reducing feature disparity, but notable performance gaps remain, particularly between controlled datasets and in-the-wild datasets. These observations indicate that aggregation alone cannot overcome supervision conflict and sample redundancy, motivating the need for mechanisms that explicitly enhance supervision consistency and sample quality in mixed-dataset gait learning.

These observations indicate that mixed-dataset training improves generalization but does not fundamentally resolve the inherent challenges introduced by heterogeneous data sources. In particular, aggregated datasets still generate inconsistent supervisory signals and contain samples of uneven reliability, which may lead to unstable optimization and suboptimal representation learning. This suggests that improving cross-domain gait performance requires not only increasing data diversity, but also explicitly enhancing supervision coherence and sample quality during training.

To address this, we develop a unified training framework that focuses on reliable supervision and stable optimization under mixed-domain data. First, we introduce a selective training strategy that reduces the influence of highly redundant or uncertain samples, enabling the model to concentrate on informative instances and mitigate bias toward noisy or domain-specific patterns. Second, we adopt a domain-separated metric formulation in which positive and negative pairs are drawn within the same dataset, preventing identity-space interference and avoiding the artificial separation of cross-domain samples during embedding learning. We further analyze Domain-Specific Batch Normalization (DSBN)~\cite{chang2019domain} in this context. DSBN improves per-domain performance by maintaining domain-dependent statistics; however, in our setup it does not yield consistent gains under cross-domain evaluation, reflecting the practical difficulty of balancing domain specialization and transferability in generalized gait learning.

Our key contributions are summarized as follows:
\begin{itemize}
    \item We analyze gait generalization under heterogeneous data distributions and identify two key challenges that hinder transferability: inconsistent supervisory signals and uneven sample reliability. This perspective clarifies why scaling training data alone does not guarantee robustness across unseen environments.
    \item We introduce a unified training framework designed to improve generalizable gait representation learning. The framework enhances supervision reliability through selective emphasis on informative samples and enforces identity-consistent metric learning to avoid cross-domain interference, enabling stable optimization and more transferable features.
    \item We conduct extensive experiments on four public gait datasets (CASIA-B~\cite{CASIA-B}, OU-MVLP~\cite{OUMVLP}, Gait3D~\cite{Gait3D}, GREW~\cite{GREW, guo2025gait}) and two representative gait architectures (GaitBase~\cite{fan2023opengait}, DeepGaitV2~\cite{fan2023exploring}). Results demonstrate consistent cross-domain improvements without compromising in-domain performance, and provide practical insights for robust gait model training under heterogeneous data.
\end{itemize}

%% file: sec/2_related_works.tex
\section{Related Works}
\label{sec:related_works}

\subsection{In-Domain Gait Recognition}
Most existing gait recognition methods are developed and evaluated in in-domain settings, where training and testing data are drawn from the same dataset. These approaches can be broadly grouped into appearance-based~\cite{yao2022improving, huang2025watch, wang2023dygait, wang2022gaitstrip}, model-based~\cite{wang2023gait}, and multi-modal methods~\cite{zhao2021associated}.
Appearance-based methods rely on silhouette sequences to model spatial-temporal gait dynamics. GaitSet~\cite{chao2019gaitset} introduces a set-based representation without explicit temporal alignment. GaitPart~\cite{fan2020gaitpart} enhances discriminability by decomposing the body into local parts, while GaitGL~\cite{lin2021gait} combines global and local feature branches. 
More recent efforts such as DeepGaitV2~\cite{fan2023exploring}, SPOSGait~\cite{guo2025gait}, and CLASH~\cite{dou2024clash} explore 3D temporal modeling and NAS-based architecture optimization.
Model-based methods utilize structured representations such as 2D/3D skeletons to extract motion cues. GaitGraph~\cite{teepe2021gaitgraph} models joint dependencies via graph convolutional networks, while SkeletonGait~\cite{fan2024skeletongait} uses heatmap encoding to enhance robustness against visual noise. In addition, several works like SkeletonGait++~\cite{fan2024skeletongait} and MultiGait++~\cite{jin2025exploring} explore multi-modal integration of silhouette, skeleton, and body-part features for improved performance under diverse conditions. However, these works often overfit to dataset-specific characteristics such as viewpoint, background, and clothing, leading to significant generalization gaps in cross-domain scenarios. This motivates the need for gait-specific strategies to handle multi-domain data composition beyond model architecture design.

\subsection{Cross-Domain Gait Recognition}
Cross-domain gait recognition~\cite{10243069} aims to build models that generalize across datasets with diverse conditions, such as varying viewpoints, clothing, and backgrounds. A key challenge lies in mitigating domain shift without access to labeled target-domain data.
Unsupervised domain adaptation (UDA) has been widely explored to align feature distributions between domains. GaitDAN~\cite{huang2024gaitdan} employs adversarial training to reduce cross-view discrepancies, while Ma \etal~\cite{ma2023fine} propose a clustering-based pseudo-labeling strategy combined with a spatio-temporal aggregation network. GPGait~\cite{fu2023gpgait} enhances pose-based adaptation via human-oriented transformation and part-aware graph convolutional learning. Trand~\cite{zheng2021trand} further improves UDA by discovering transferable local neighborhoods in the embedding space.
Domain generalization approaches seek to improve robustness without using target-domain samples. CDTN~\cite{tong2019gait} transfers latent representations through cross-domain mappings. BigGait~\cite{ye2024biggait} introduces a pipeline that leverages large vision models and a Gait Representation Extractor (GRE) to produce task-relevant features from generic embeddings, achieving strong performance in cross-domain evaluation. In parallel, self-supervised learning methods such as GaitSSB~\cite{fan2023learning} exploit unlabeled sequences to extract transferable representations, but they often struggle to maintain discriminative identity cues across heterogeneous domains. Jaiswal \etal~\cite{jaiswal2024domain} explore domain-specific adaptation modules designed for practical deployment under unknown conditions.

Despite these advances, most existing methods either depend on target-domain statistics or introduce auxiliary adaptation modules, while overlooking the challenges introduced by mixed-dataset training, such as inter-dataset optimization conflicts and data redundancy. These issues remain largely underexplored in current literature and motivate the need for more scalable, unified solutions.

\subsection{Dataset Distillation}
Dataset distillation aim to improve training efficiency by identifying and retaining only informative subsets of data. In image classification, early works have explored synthetic distillation~\cite{wang2018dataset}, soft-label-based selection~\cite{sucholutsky2021soft}, and latent factorization for data compression~\cite{liu2022dataset}. More recent approaches improve distillation realism and diversity via patch recomposition and retrieval-based strategies~\cite{sun2024diversity}. In the context of human analysis, data pruning and filtering techniques have been applied to face recognition~\cite{schlett2024double}, re-identification~\cite{yao2023large}, and video understanding~\cite{gowda2021smart}, where redundant or low-quality samples may impair generalization. 

However, most existing methods operate in single-domain settings with relatively homogeneous data distributions. By contrast, gait data span controlled laboratory and in-the-wild environments, leading to substantial domain heterogeneity, silhouette noise, and viewpoint–motion coupling. These characteristics make redundancy and noise more domain-specific and amplify their impact during mixed-dataset training. Our work investigates sample selection under this heterogeneous gait setting, aiming to preserve informative sequences while suppressing redundant or noisy ones.

%% file: sec/3_method.tex
\section{Method}
\label{sec:method}

\begin{figure*}[t]
  \centering
  \includegraphics[width=0.98\linewidth]{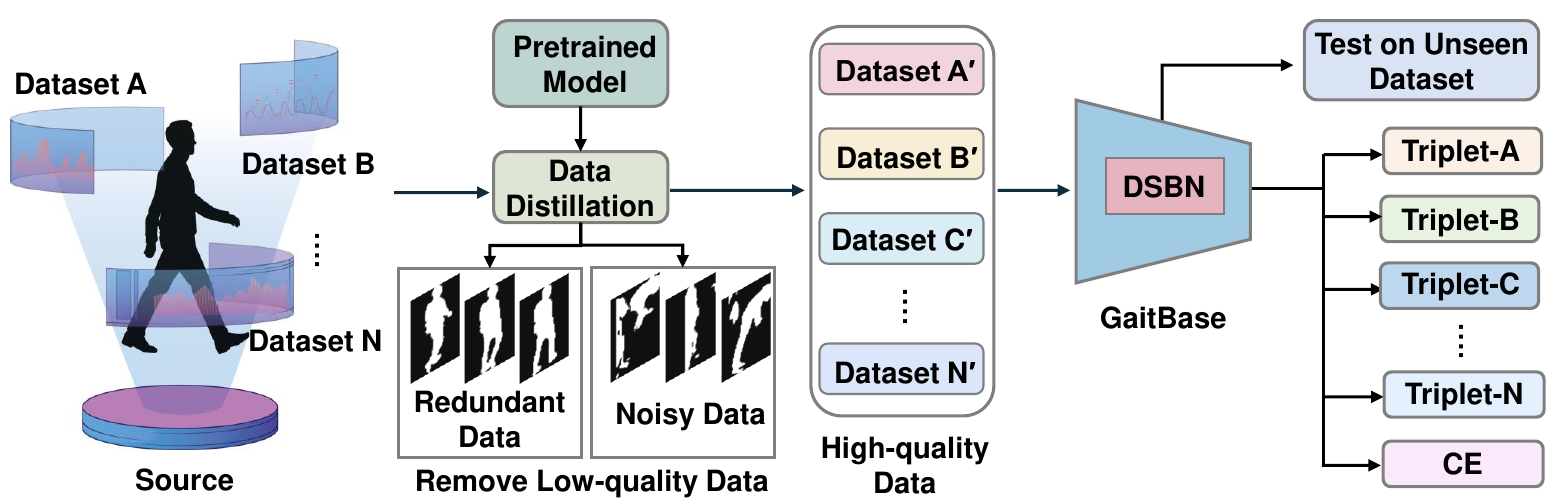}%
  \caption{Overall framework of the proposed approach for cross-domain gait recognition. Each source dataset undergoes dataset distillation using a pre-trained model to filter out redundant and noisy samples, yielding high-quality subsets. These subsets are then combined to train the GaitBase~\cite{fan2023opengait} model with Domain-Specific Batch Normalization (DSBN)~\cite{chang2019domain} and separate triplet losses, enhancing cross-dataset performance.}
  \label{fig:main}
\end{figure*}

\subsection{Dataset Distillation for Cross-Domain Robustness}
While aggregating multiple datasets increases sample diversity, it also introduces domain-specific biases. Controlled indoor datasets often contain highly repetitive walking sequences, whereas outdoor datasets include occlusions, clutter, and segmentation noise. Directly mixing all samples may therefore lead the model to emphasize trivial indoor patterns or be affected by unreliable outdoor observations, making it difficult to learn robust, domain-invariant gait representations.

To address this issue, we introduce a dataset distillation strategy that selectively removes uninformative samples before mixing datasets. For each dataset $\mathcal{D}_k$, we use a pretrained GaitBase~\cite{fan2023opengait} model (trained only on $\mathcal{D}_k$) to extract features $F_i = f_\theta(X_i)$ and compute reliability metrics in the feature space. This distillation stage retains representative and reliable samples, forming a compact and clean subset that preserves identity diversity while suppressing redundancy and annotation noise. Intuitively, samples that are either too easy or inherently unreliable contribute little stable learning signal; filtering them serves as a data-centric regularization mechanism, improving the efficiency and robustness of multi-dataset training.

\subsubsection{Reducing Redundancy in Indoor Datasets}
Indoor gait datasets are typically collected in controlled environments with consistent backgrounds, viewpoints, and walking trajectories. As a result, they often contain a large number of highly similar samples that provide limited additional variation. Such redundant samples may lead the model to memorize trivial appearance cues rather than learning generalizable gait dynamics.

To suppress redundancy, we measure the contribution of each sample by computing its average Euclidean distance to negative samples (samples from different identities) in the feature space:
\begin{equation}
    \text{mean\_dist}(X_i)
    = \frac{1}{|N_i|} \sum_{X_j \in N_i} D(F_i, F_j)
\end{equation}
where $F_i = f_\theta(X_i)$ denotes the embedding of sample $X_i$, $N_i$ is the set of negative samples, and $D(\cdot,\cdot)$ is the Euclidean distance. A large $\text{mean\_dist}(X_i)$ indicates that $X_i$ already lies far from all negatives, meaning its margin to other identities is well satisfied.

Intuitively, such samples rarely participate in margin-violating triplets and thus contribute negligible gradient updates during metric learning. Removing them preserves the effective decision boundary while reducing redundancy, allowing the model to focus on informative and boundary-critical samples that better promote discriminative and compact representations.

\subsubsection{Removing Noisy Samples in Outdoor Datasets}
Outdoor gait datasets are affected by uncontrolled factors such as occlusions, moving backgrounds, illumination changes, and imperfect silhouette extraction. These issues produce noisy or distorted samples that deviate from typical gait patterns and may mislead the learning process by introducing unstable identity cues.

To detect such unreliable samples, we first measure the intra-class consistency of each sample by computing its distance to the identity centroid in the feature space:
\begin{equation}
    \mu_k = \frac{1}{|\mathcal{C}_k|} \sum_{X_j \in \mathcal{C}_k} F_j
\end{equation}
\begin{equation}
    \text{intra\_dist}(X_i) = D(F_i, \mu_k)
\end{equation}
where $F_i = f_\theta(X_i)$ denotes the embedding of $X_i$ and $\mathcal{C}_k$ denotes the set of samples belonging to identity $k$. Samples with abnormally large $\text{intra\_dist}$ values are assumed to be noisy or corrupted, as they diverge from the identity cluster and are less likely to represent stable gait structure.

In addition, we use part-level prediction consistency as a complementary reliability cue. Following BNNeck~\cite{BNNecks}, each gait sequence is divided into $p$ horizontal parts via Horizontal Pyramid Pooling (HPP)~\cite{HPP}, and the model predicts identity for each part. Conceptually, clean gait sequences exhibit coherent motion patterns across body regions, thus producing consistent identity predictions. In contrast, samples affected by occlusion, segmentation artifacts, or missing body parts typically corrupt only certain regions (e.g., legs covered by obstacles), resulting in inconsistent part-level predictions. We therefore mark a sample as unreliable if any part-level prediction differs from the ground truth:
\begin{equation}
    \text{failure}_i =
    \begin{cases}
        1, & \text{if} \bigvee_{j=1}^{p} \left( \text{preds}_{i,j} \neq \text{labels}_i \right) \\[8pt]
        0, & \text{otherwise}.
    \end{cases}
\end{equation}
This strategy targets structurally corrupted samples rather than genuinely hard examples, preventing unstable gradients caused by noisy partial silhouettes while retaining challenging but valid training instances.

We remove the top $n\%$ of samples with large intra-class distances or frequent prediction failures. This procedure eliminates ambiguous and noisy observations, stabilizing the identity manifold and ensuring that the remaining samples provide consistent and reliable supervision for cross-domain learning.

\subsection{Separate Triplet Loss for Multi-Dataset Training}
Triplet loss is widely adopted in gait recognition to enforce compact intra-class embeddings while enlarging inter-class margins. 
However, when training with multiple datasets simultaneously, directly treating all samples outside the anchor's identity as negatives introduces domain-level interference. 
Since identity labels are disjoint across datasets, cross-domain samples are always regarded as negatives, which encourages the model to separate domains rather than individuals. 
This leads the network to unintentionally capture dataset-specific cues (e.g., background style, silhouette distribution) instead of identity-related gait patterns, weakening cross-domain generalization.

From a mathematical perspective, consider the triplet hinge objective. Let $F_i$, $F_j$, and $F_k$ denote the embeddings of the anchor, positive, and negative samples, respectively:
\begin{equation}
\ell = [D(F_i,F_j) - D(F_i,F_k) + m]_+ 
\end{equation}
For cross-domain negative pairs, distribution shift often yields 
\begin{equation}
D(F_i,F_k) \gg D(F_i,F_j)
\end{equation}
even when the two identities exhibit similar gait motion. Such cases are incorrectly interpreted as hard negatives, activating the hinge term and producing gradients that further enlarge inter-domain gaps rather than identity margins. This domain-induced repulsion gradually distorts the embedding space, pushing domains apart and degrading performance on unseen datasets.

To avoid such conflict, we compute triplet loss independently within each dataset, restricting anchor, positive, and negative samples to originate from the same domain. 
For dataset $\mathcal{D}_k$, the loss is defined as:
\begin{equation}
    \text{Loss}_{\text{tri}}^{k}
    = \left[ D(F_i, F_j) - D(F_i, F_k) + m \right]_{+}
\end{equation}
where $D(\cdot)$ denotes Euclidean distance, $m$ is the margin, and $[\cdot]_+$ is the ReLU function. 
This intra-domain formulation preserves identity discrimination while preventing dataset-specific distribution gaps from distorting the metric space.

The overall objective combines the dataset-specific triplet losses with a unified cross-entropy identity loss:
\begin{equation}
    \text{Loss}_{\text{all}}
    = \sum_{k=1}^{n} w^k \text{Loss}_{\text{tri}}^{k}
    + \text{Loss}_{\text{ce}}
\end{equation}
where $w^k$ balances contributions from each dataset. By isolating metric learning within each domain, the proposed loss formulation prevents artificial domain separation, stabilizes optimization, and encourages learning domain-invariant gait representations that generalize effectively to unseen datasets.

\subsection{Domain-Specific Batch Normalization}
Batch Normalization (BN) tightly couples feature statistics with data distribution. When multiple datasets exhibit distinct appearance and silhouette characteristics, a single BN layer may mix heterogeneous statistics and introduce domain bias. To address this, we incorporate Domain-Specific Batch Normalization (DSBN)~\cite{chang2019domain} in the mixed-dataset training stage. For each dataset $\mathcal{D}_k$, an independent BN branch maintains its own mean, variance, and affine parameters $(\mu_k,\sigma_k^2,\gamma_k,\beta_k)$:
\begin{equation}
    \text{BN}_k(x)=\gamma_k \frac{x-\mu_k}{\sqrt{\sigma_k^2 + \epsilon}} + \beta_k
\end{equation}
This design enables dataset-specific normalization, preventing cross-domain interference and allowing the shared backbone to focus on learning domain-invariant gait structure.

During inference on an unseen dataset where domain-specific BN parameters are not available, we adopt an output-averaging strategy. 
Rather than estimating new statistics or selecting a single domain branch, we apply all BN branches to the input and average their normalized activations:
\begin{equation}
    \text{BN}_{\text{avg}}(x)=\frac{1}{n}\sum_{k=1}^{n}\text{BN}_k(x)
\end{equation}
where $n$ is the number of training domains. 
This approach implicitly aggregates learned domain priors and provides a stable normalization for unseen domains without requiring domain labels or additional calibration.

%% file: sec/4_experiments.tex
\section{Experiments}
\label{sec:experiments}

\subsection{Datasets and Metrics}
We evaluate our approach on four widely used gait datasets: CASIA-B~\cite{CASIA-B}, OU-MVLP~\cite{OUMVLP}, Gait3D~\cite{Gait3D}, and GREW~\cite{GREW}.

\textbf{CASIA-B}~\cite{CASIA-B} is an indoor dataset with 124 subjects captured from 11 viewpoints ($0^\circ$ to $180^\circ$, at $18^\circ$ intervals) under three conditions: normal walking (NM), walking with a bag (BG), and walking in a coat (CL). We follow the standard protocol~\cite{chao2019gaitset}, using 74 subjects for training and 50 for testing. NM\#01-04 serve as the gallery, while NM\#05-06, BG\#01-02, and CL\#01-02 form the probe set.

\begin{table}[ht]
    \centering
    \caption{Training schedule across different datasets (steps in thousands).}
    \begin{tabular}{ccc}
        \toprule
        Dataset & Decay Steps (k) & Total Steps (k) \\
        \midrule
        CASIA-B & (20, 40, 60) & 80 \\
        OU-MVLP & (30, 60, 90) & 120 \\
        Gait3D & (20, 40, 60) & 80 \\
        GREW & (40, 80, 120) & 160 \\
        \midrule
        CASIA-B, OUMVLP, Gait3D & (30, 60, 90) & 120 \\
        CASIA-B, OUMVLP, GREW & (40, 80, 120) & 160 \\
        CASIA-B, Gait3D, GREW & (40, 80, 120) & 160 \\
        OUMVLP, Gait3D, GREW & (40, 80, 120) & 160 \\
        \bottomrule
    \end{tabular}
    \label{tab:training_details}
\end{table}

\begin{table*}[t]
    \centering
    \caption{Cross-dataset validation and mixed dataset training results for GaitBase~\cite{fan2023opengait} and DeepGaitV2~\cite{fan2023exploring}. * denotes the distilled subset (20\% data removal). Self-domain results are highlighted in \colorbox{selfdomain}{light green}, and cross-domain results in \colorbox{crossdomain}{light yellow}. Best and suboptimal results are marked in \textbf{bold} and \underline{underlined}, respectively.}
    \label{main_results:combined}
    \begin{tabular}{c|c|c|c|c|c|c|c|c}
        \toprule
        \multirow{2}{*}{Training Set} & \multicolumn{4}{c|}{GaitBase~\cite{fan2023opengait}} & \multicolumn{4}{c}{DeepGaitV2~\cite{fan2023exploring}} \\
        \cline{2-9}
        & CASIA-B & OUMVLP & Gait3D & GREW & CASIA-B & OUMVLP & Gait3D & GREW \\
        \midrule
        CASIA-B & \cellcolor{selfdomain}88.88 & \cellcolor{crossdomain}24.64 & \cellcolor{crossdomain}15.30 & \cellcolor{crossdomain}17.43 & \cellcolor{selfdomain}89.69 & \cellcolor{crossdomain}29.82 & \cellcolor{crossdomain}15.20 & \cellcolor{crossdomain}19.78 \\
        OU-MVLP & \cellcolor{crossdomain}61.91 & \cellcolor{selfdomain}\textbf{89.23} & \cellcolor{crossdomain}19.20 & \cellcolor{crossdomain}23.18 & \cellcolor{crossdomain}69.70 & \cellcolor{selfdomain}\textbf{91.93} & \cellcolor{crossdomain}25.20 & \cellcolor{crossdomain}32.55 \\
        Gait3D & \cellcolor{crossdomain}53.21 & \cellcolor{crossdomain}39.99 & \cellcolor{selfdomain}63.10 & \cellcolor{crossdomain}30.22 & \cellcolor{crossdomain}53.96 & \cellcolor{crossdomain}44.73 & \cellcolor{selfdomain}76.60 & \cellcolor{crossdomain}35.20 \\
        GREW & \cellcolor{crossdomain}49.50 & \cellcolor{crossdomain}30.36 & \cellcolor{crossdomain}27.10 & \cellcolor{selfdomain}\textbf{58.16} & \cellcolor{crossdomain}51.96 & \cellcolor{crossdomain}36.62 & \cellcolor{crossdomain}32.40 & \cellcolor{selfdomain}\textbf{75.31} \\
        \midrule
        CASIA-B, OUMVLP, Gait3D & \cellcolor{selfdomain}\underline{90.90} & \cellcolor{selfdomain}84.08 & \cellcolor{selfdomain}62.40 & \cellcolor{crossdomain}\underline{32.68} & \cellcolor{selfdomain}\underline{91.45} & \cellcolor{selfdomain}87.96 & \cellcolor{selfdomain}76.20 & \cellcolor{crossdomain}\underline{36.70} \\
        CASIA-B$^*$, OUMVLP$^*$, Gait3D$^*$ & \cellcolor{selfdomain}89.70 & \cellcolor{selfdomain}83.83 & \cellcolor{selfdomain}62.10 & \cellcolor{crossdomain}\textbf{32.90} & \cellcolor{selfdomain}90.89 & \cellcolor{selfdomain}87.55 & \cellcolor{selfdomain}75.80 & \cellcolor{crossdomain}\textbf{37.00} \\
        \midrule
        CASIA-B, OUMVLP, GREW & \cellcolor{selfdomain}\textbf{91.28} & \cellcolor{selfdomain}85.22 & \cellcolor{crossdomain}\underline{29.70} & \cellcolor{selfdomain}57.05 & \cellcolor{selfdomain}\textbf{91.98} & \cellcolor{selfdomain}88.12 & \cellcolor{crossdomain}\underline{35.10} & \cellcolor{selfdomain}74.68 \\
        CASIA-B$^*$, OUMVLP$^*$, GREW$^*$ & \cellcolor{selfdomain}90.79 & \cellcolor{selfdomain}\underline{85.32} & \cellcolor{crossdomain}\textbf{30.30} & \cellcolor{selfdomain}56.74 & \cellcolor{selfdomain}91.62 & \cellcolor{selfdomain}\underline{87.80} & \cellcolor{crossdomain}\textbf{35.60} & \cellcolor{selfdomain}74.35 \\
        \midrule
        CASIA-B, Gait3D, GREW & \cellcolor{selfdomain}87.60 & \cellcolor{crossdomain}\underline{43.22} & \cellcolor{selfdomain}\underline{63.60} & \cellcolor{selfdomain}56.34 & \cellcolor{selfdomain}88.55 & \cellcolor{crossdomain}\underline{48.60} & \cellcolor{selfdomain}\underline{76.80} & \cellcolor{selfdomain}74.15 \\
        CASIA-B$^*$, Gait3D$^*$, GREW$^*$ & \cellcolor{selfdomain}87.49 & \cellcolor{crossdomain}\textbf{44.58} & \cellcolor{selfdomain}\textbf{64.00} & \cellcolor{selfdomain}\underline{57.06} & \cellcolor{selfdomain}88.40 & \cellcolor{crossdomain}\textbf{49.30} & \cellcolor{selfdomain}\textbf{77.20} & \cellcolor{selfdomain}\underline{74.50} \\
        \midrule
         OUMVLP, Gait3D, GREW & \cellcolor{crossdomain}\textbf{64.09} & \cellcolor{selfdomain}83.75 & \cellcolor{selfdomain}60.60 & \cellcolor{selfdomain}55.63 & \cellcolor{crossdomain}\textbf{66.50} & \cellcolor{selfdomain}85.45 & \cellcolor{selfdomain}72.10 & \cellcolor{selfdomain}72.85 \\
        OUMVLP$^*$, Gait3D$^*$, GREW$^*$ & \cellcolor{crossdomain}\underline{63.72} & \cellcolor{selfdomain}82.94 & \cellcolor{selfdomain}61.10 & \cellcolor{selfdomain}55.34 & \cellcolor{crossdomain}\underline{65.93} & \cellcolor{selfdomain}84.92 & \cellcolor{selfdomain}72.70 & \cellcolor{selfdomain}72.50 \\
        \midrule
        CASIA-B, OUMVLP, Gait3D, GREW & \cellcolor{selfdomain}88.20 & \cellcolor{selfdomain}84.14 & \cellcolor{selfdomain}58.50 & \cellcolor{selfdomain}54.13 & \cellcolor{selfdomain}89.24 & \cellcolor{selfdomain}85.16 & \cellcolor{selfdomain}76.20 & \cellcolor{selfdomain}73.81 \\
        CASIA-B$^*$, OUMVLP$^*$, Gai3D$^*$, GREW$^*$ & \cellcolor{selfdomain}88.42 & \cellcolor{selfdomain}84.70 & \cellcolor{selfdomain}58.90 & \cellcolor{selfdomain}54.42 & \cellcolor{selfdomain}89.68 & \cellcolor{selfdomain}85.73 & \cellcolor{selfdomain}75.90 & \cellcolor{selfdomain}73.43 \\
        \bottomrule
    \end{tabular}
\end{table*}

\textbf{OU-MVLP}~\cite{OUMVLP} is a large-scale dataset with 10,307 subjects captured from 14 viewpoints ($0^\circ$ to $90^\circ$ and $180^\circ$ to $270^\circ$, at $15^\circ$ intervals). We use 5,153 subjects for training and the rest for testing, where Seq\#01 is the gallery and Seq\#00 is the probe.

\textbf{Gait3D}~\cite{Gait3D} is an unconstrained dataset collected in an indoor supermarket with 39 cameras, containing 4,000 subjects and over 25,000 sequences. The training set includes 3,000 subjects, while 1,000 subjects are used for testing. One sequence per subject serves as the probe, with the rest forming the gallery.

\textbf{GREW}~\cite{GREW} is a large-scale outdoor dataset with 26,345 subjects captured by 882 cameras in real-world environments. It comprises 128,671 sequences, split into a training set (20,000 subjects), validation set (345 subjects), and test set (6,000 subjects). Each test subject has four sequences, with two assigned as the gallery and two as the probe.

We use Rank-1 accuracy as the primary evaluation metric to compare recognition performance across datasets.

\subsection{Implementation Details}
All experiments are conducted using the PyTorch framework on \(8 \times\) NVIDIA RTX 4090 GPUs, with GaitBase~\cite{fan2023opengait} or DeepGaitv2~\cite{fan2023exploring} as the backbone. The input resolution is fixed at \(64 \times 44\) pixels, and data augmentation techniques such as Random Perspective Transformation, Horizontal Flipping, and Random Rotation are applied. The model is trained with SGD, using an initial learning rate of 0.1, a weight decay of $5e-4$, and a momentum of 0.9. The margin $m$ in triplet loss is set to 0.2, following the GaitBase~\cite{fan2023opengait} optimization strategy.

To enhance generalization, we adopt a mixed dataset sampling strategy, ensuring balanced representation from multiple datasets. For each dataset $\mathcal{D}_i$, $P_i$ identities and $K_i$ sequences per identity are randomly sampled, forming a mini-batch of size $B=\sum_{i=1}^{n} P_i \times K_i$. The dataset-specific batch sizes are: (\(16,4\)) for CASIA-B~\cite{CASIA-B}, and (\(32,4\)) for OU-MVLP~\cite{OUMVLP}, GREW~\cite{GREW}, and Gait3D~\cite{Gait3D}. In mixed training, triplet loss weights $w^k$ are set as: 0.2 for CASIA-B~\cite{CASIA-B}, 0.4 for OU-MVLP~\cite{OUMVLP}, 1.0 for GREW~\cite{GREW}, and 0.8 for Gait3D~\cite{Gait3D}. Learning rate schedules follow a multi-step decay, detailed in Table~\ref{tab:training_details}.

\subsection{Mixed Dataset Results}
\label{sec:mixed_dataset_results}
We present cross-dataset and mixed-dataset training results for both GaitBase~\cite{fan2023opengait} and DeepGaitV2~\cite{fan2023exploring} in Table~\ref{main_results:combined}. Mixed-dataset training, which incorporates both in-the-lab and in-the-wild datasets, not only leads to moderate improvements in self-domain accuracy but also brings substantial gains in cross-domain generalization for both models. This demonstrates that models can benefit from the complementary properties of different datasets: indoor datasets such as CASIA-B~\cite{CASIA-B} provide clean silhouettes and balanced viewpoints, while large-scale outdoor datasets like GREW~\cite{GREW} and Gait3D~\cite{Gait3D} introduce richer scene variations, complex backgrounds, and diverse subject appearances. By exposing models to such heterogeneous data distributions, mixed training encourages the extraction of more robust gait representations that remain effective under domain shift.

For example, integrating OU-MVLP~\cite{OUMVLP}, CASIA-B~\cite{CASIA-B}, and Gait3D~\cite{Gait3D} datasets achieves self-domain accuracies of 90.90\% and 91.45\% on CASIA-B~\cite{CASIA-B} using GaitBase~\cite{fan2023opengait} and DeepGaitV2~\cite{fan2023exploring}, respectively. These results show that combining large-scale multi-view data from OU-MVLP with the scene diversity of Gait3D helps reinforce recognition performance even in controlled indoor settings. Similarly, combining CASIA-B~\cite{CASIA-B}, OU-MVLP~\cite{OUMVLP}, and GREW~\cite{GREW} datasets yields the best self-domain accuracy of 91.98\% on CASIA-B~\cite{CASIA-B} using DeepGaitV2~\cite{fan2023exploring}, highlighting that viewpoint diversity from OU-MVLP enhances robustness and that outdoor variability from GREW further regularizes the learned features.

In cross-domain evaluations, mixed-dataset training consistently outperforms single-dataset training and shows clear advantages when transferring across challenging environments. Specifically, training DeepGaitV2~\cite{fan2023exploring} on CASIA-B~\cite{CASIA-B}, OU-MVLP~\cite{OUMVLP}, and GREW~\cite{GREW} datasets significantly improves accuracy to 35.10\% on Gait3D~\cite{Gait3D}, surpassing the single-dataset baseline by nearly 10\%. This suggests that including both indoor and outdoor datasets reduces overfitting to a single distribution and improves adaptability to unseen, unconstrained scenarios. Likewise, DeepGaitV2 achieves a remarkable cross-domain accuracy of 66.50\% on CASIA-B~\cite{CASIA-B} when trained jointly on OU-MVLP~\cite{OUMVLP}, Gait3D~\cite{Gait3D}, and GREW~\cite{GREW}, substantially outperforming individual dataset training. These results indicate that incorporating multi-source variability—viewpoints, camera setups, and environmental complexity—greatly enhances generalization capability, making mixed-dataset training a highly effective strategy for cross-domain gait recognition.

% These results demonstrate that exposure to diverse gait distributions significantly enhances model robustness, making mixed-dataset training a promising approach for cross-domain gait recognition.

\subsection{Dataset Distillation Results}
\label{sec:distillation_results}

\begin{figure*}[t]
  \centering
  \captionsetup[subfloat]{labelformat=empty} % 取消上方子图的(a)(b)(c)(d)编号
  % 第一行 (GaitBase)
  \subfloat[]{%
    \includegraphics[width=0.24\linewidth]{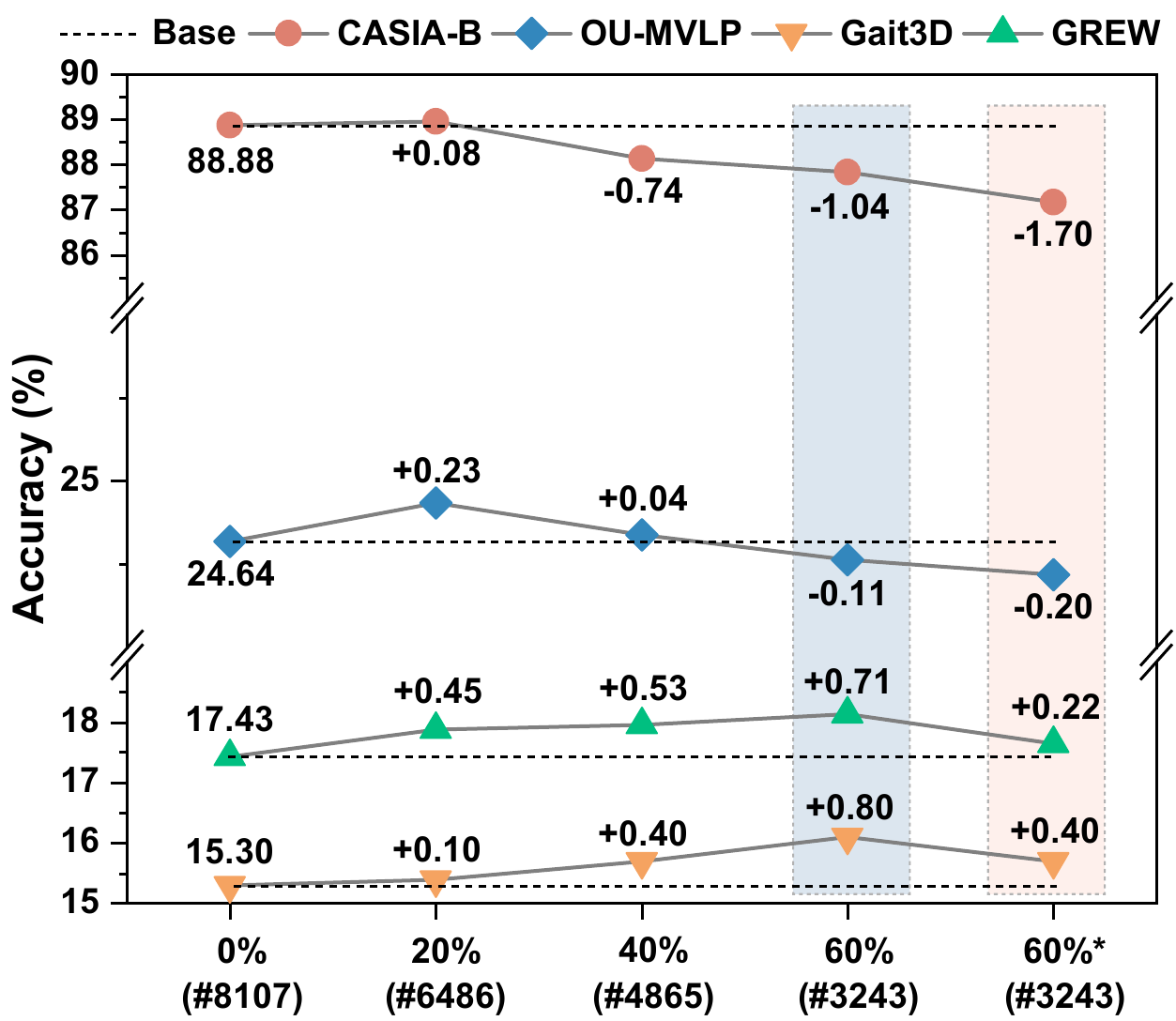}%
    \label{fig:gaitbase_casiab}%
  }
  \subfloat[]{%
    \includegraphics[width=0.24\linewidth]{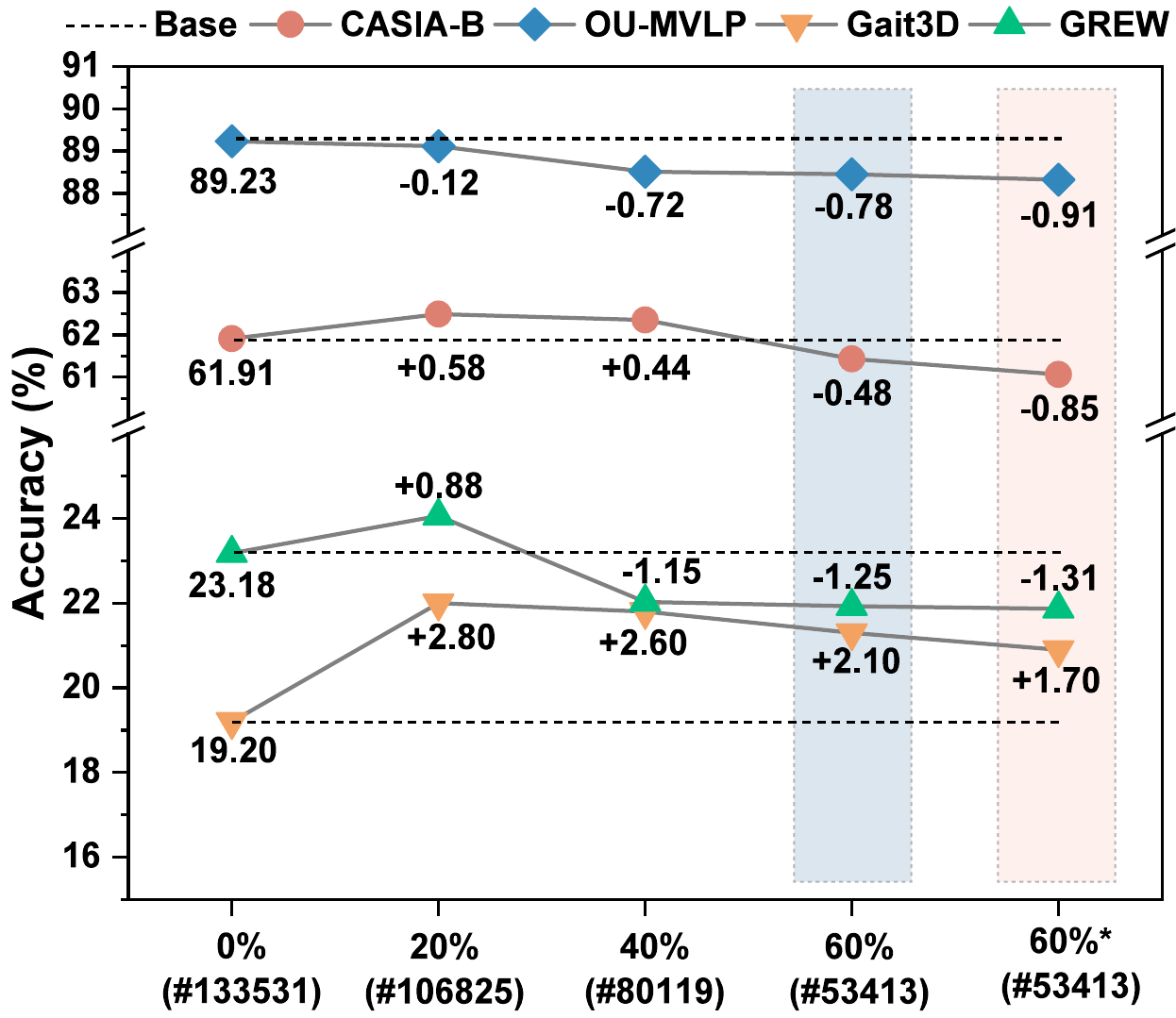}%
    \label{fig:gaitbase_oumv}%
  } 
  \subfloat[]{%
    \includegraphics[width=0.24\linewidth]{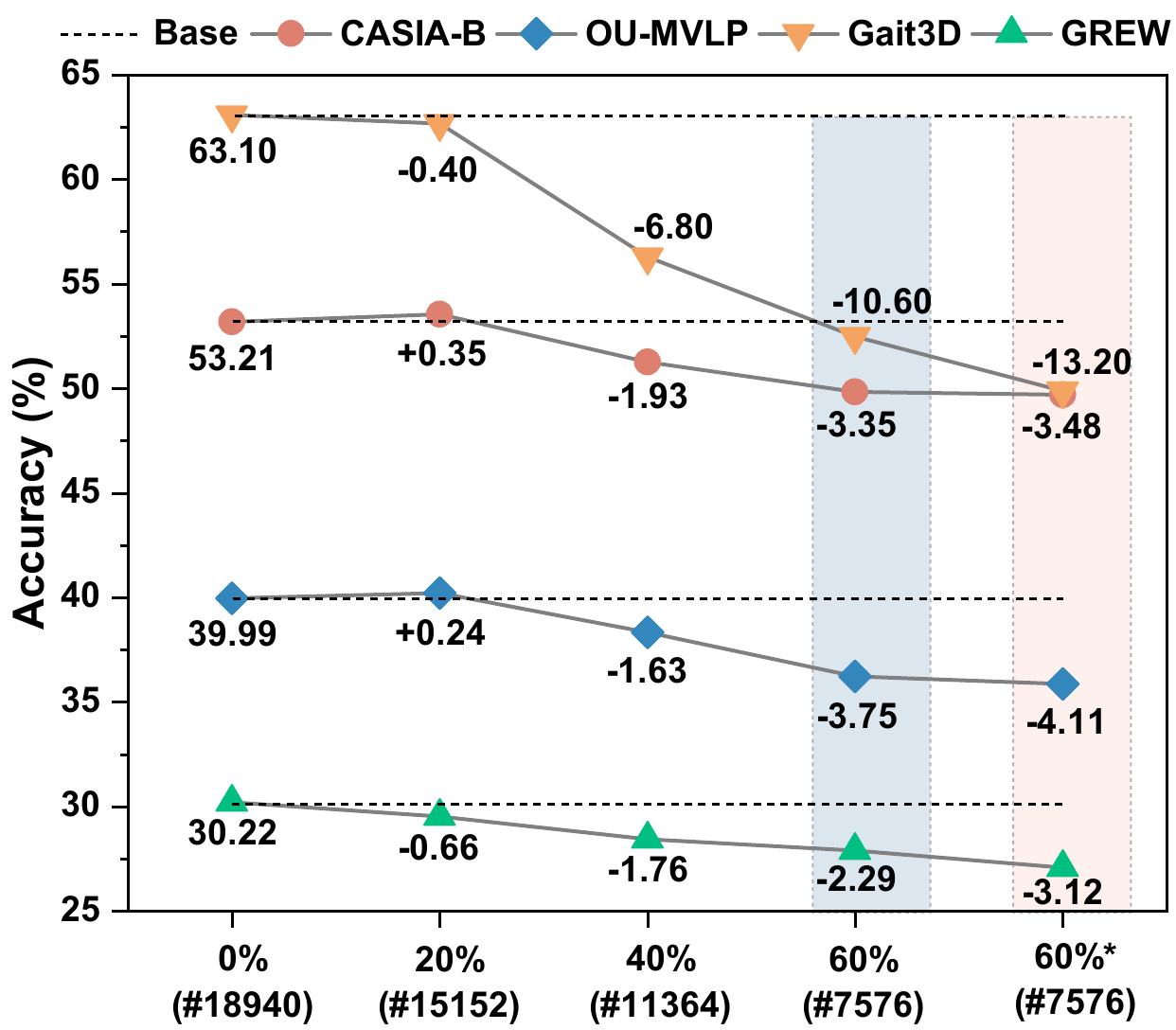}%
    \label{fig:gaitbase_gait3d}%
  }
  \subfloat[]{%
    \includegraphics[width=0.24\linewidth]{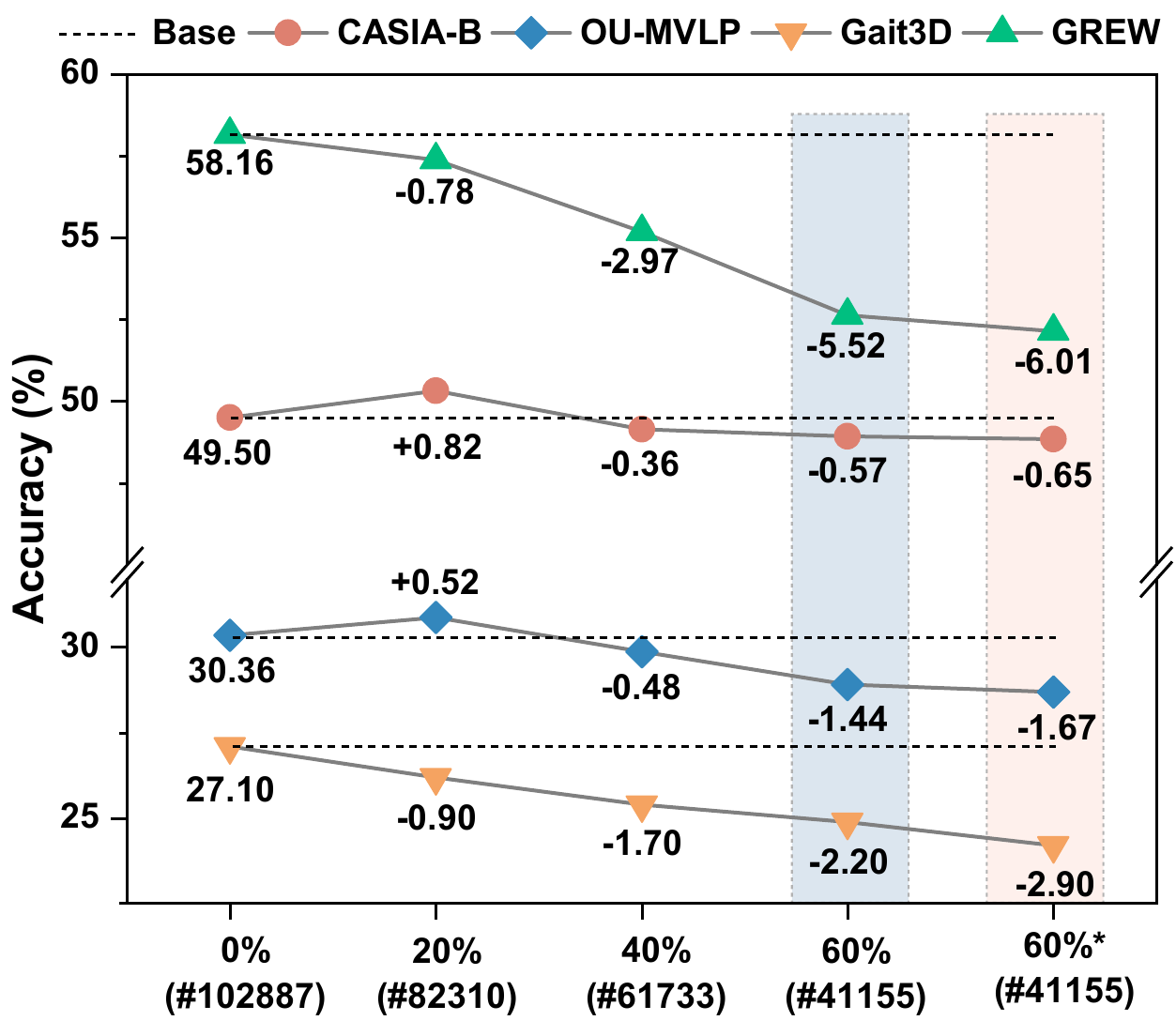}%
    \label{fig:gaitbase_grew}%
  }\\ % 换行
  
  % 第二行 (DeepGaitV2)
  \subfloat[(a) CASIA-B~\cite{CASIA-B} subsets]{%
    \includegraphics[width=0.24\linewidth]{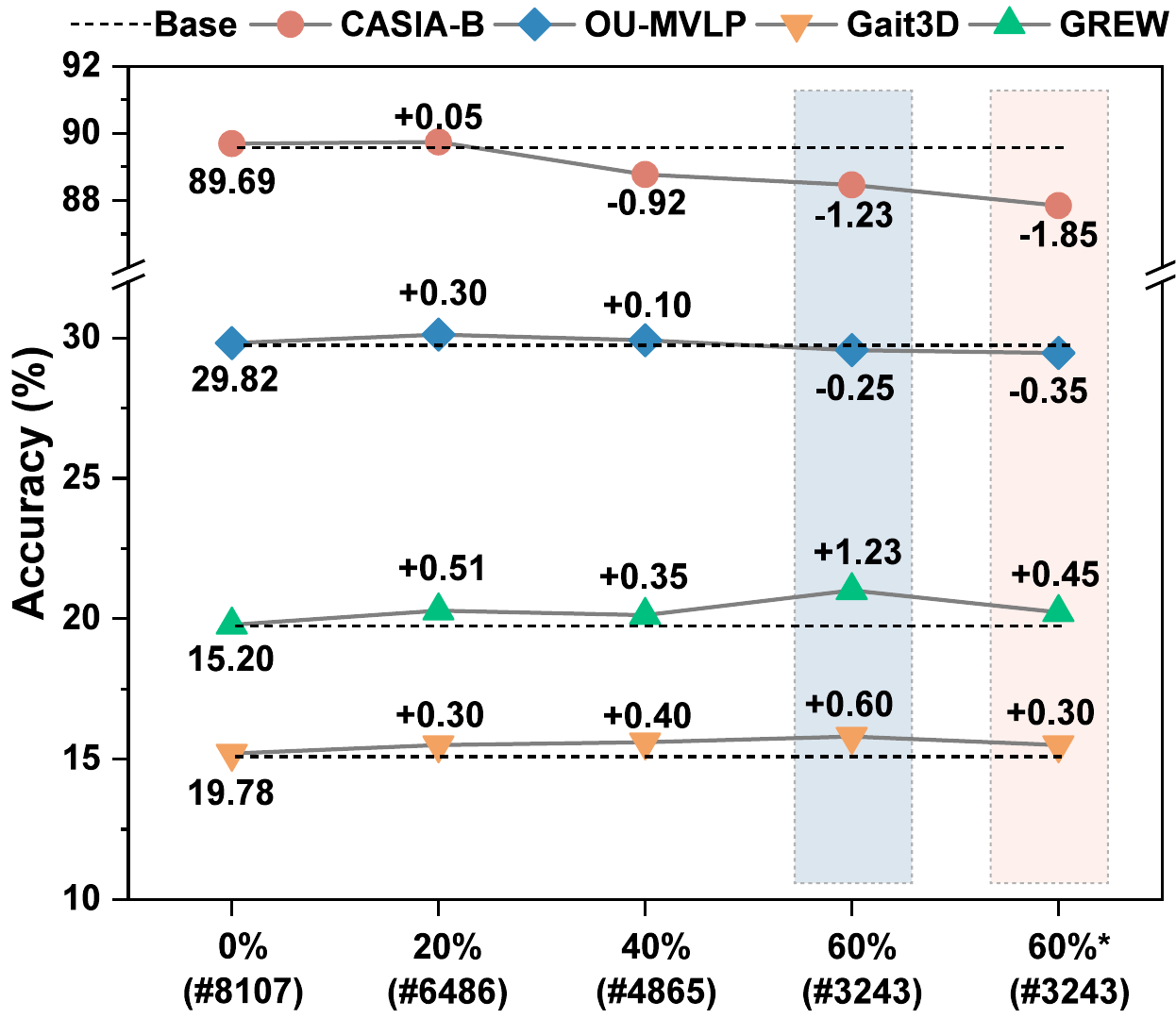}%
    \label{fig:deepgaitv2_casiab}%
  }
  \subfloat[(b) OU-MVLP~\cite{OUMVLP} subsets]{%
    \includegraphics[width=0.24\linewidth]{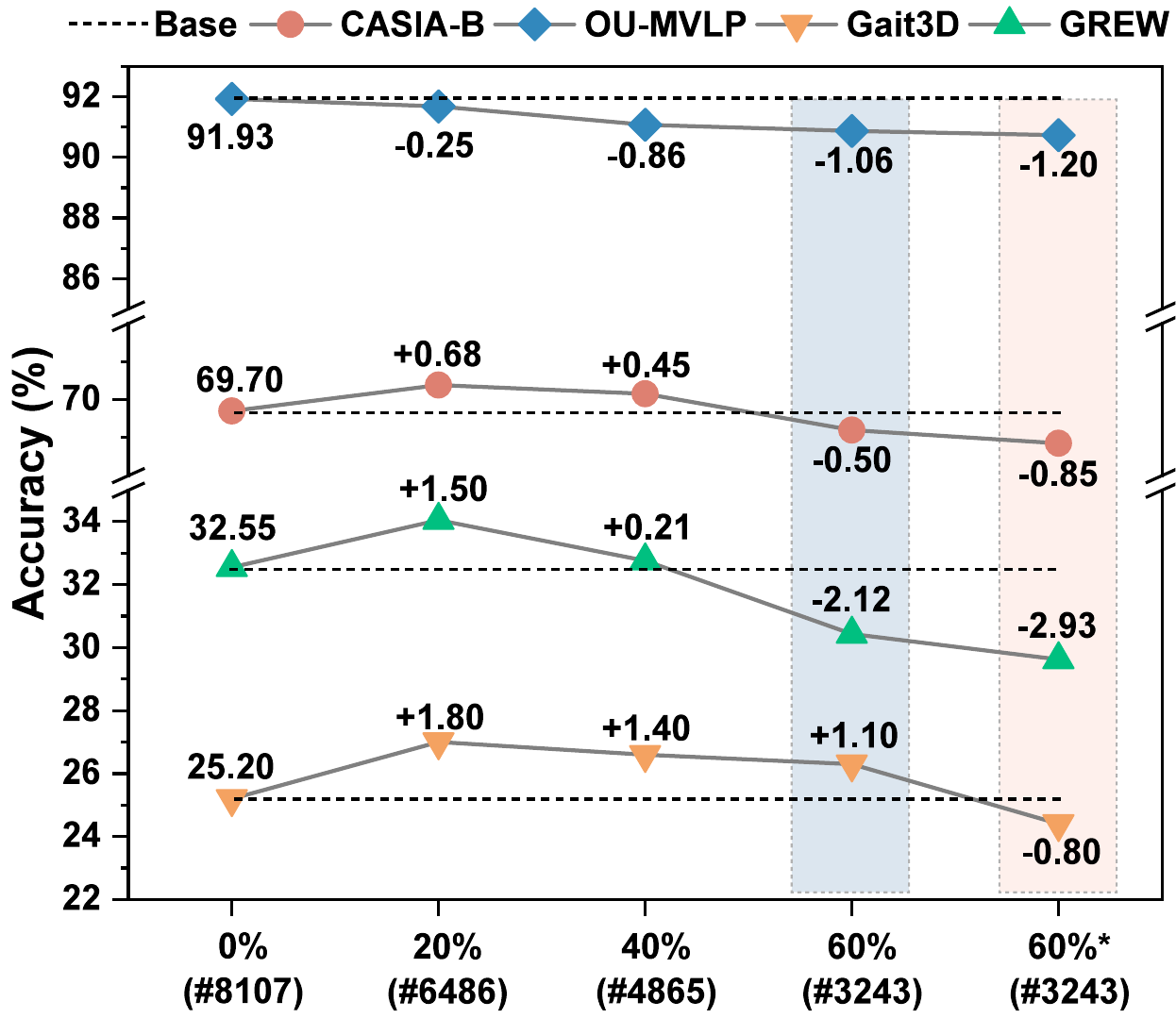}%
    \label{fig:deepgaitv2_oumv}%
  } 
  \subfloat[(c) Gait3D~\cite{Gait3D} subsets]{%
    \includegraphics[width=0.24\linewidth]{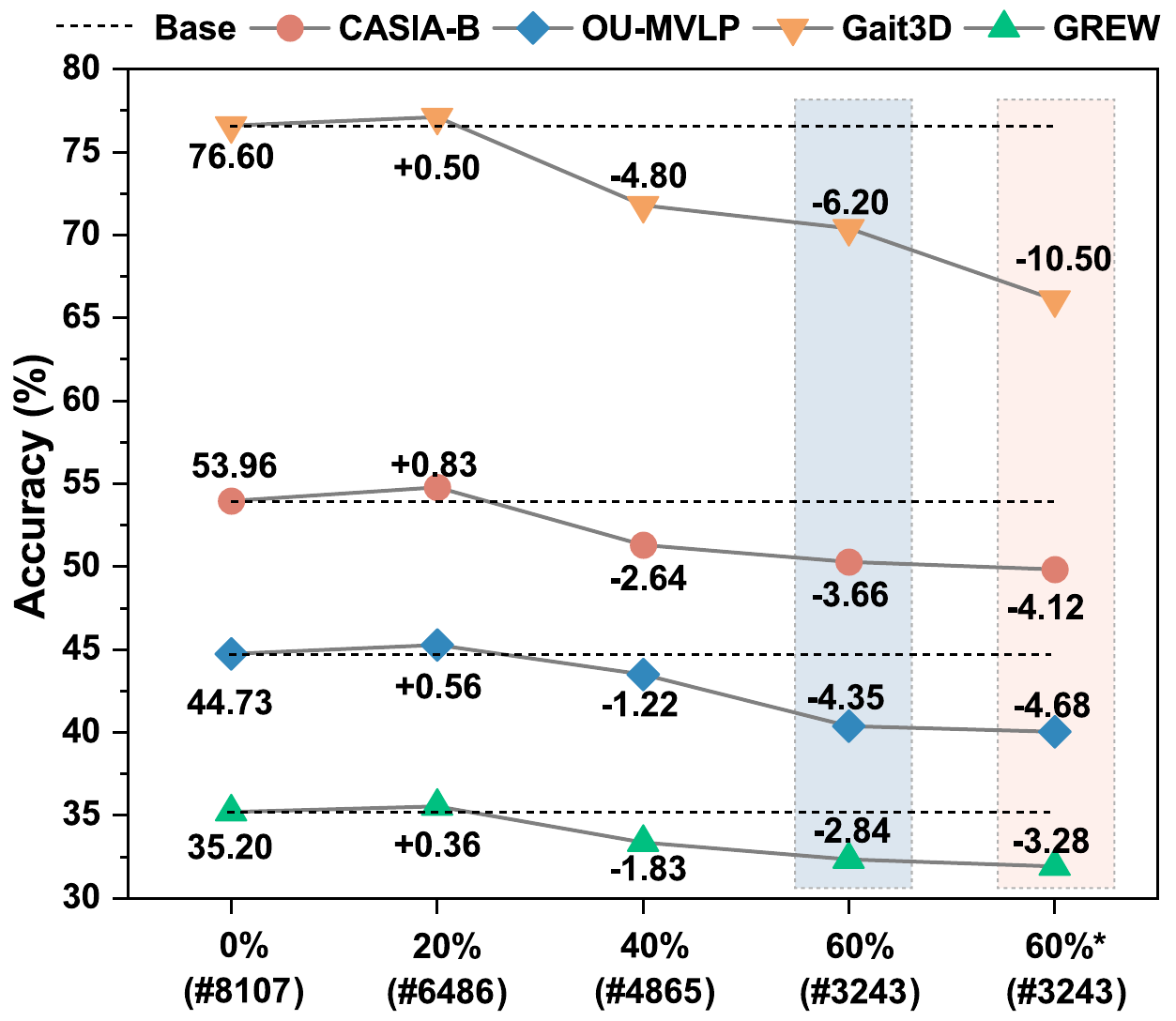}%
    \label{fig:deepgaitv2_gait3d}%
  }
  \subfloat[(d) GREW~\cite{GREW} subsets]{%
    \includegraphics[width=0.24\linewidth]{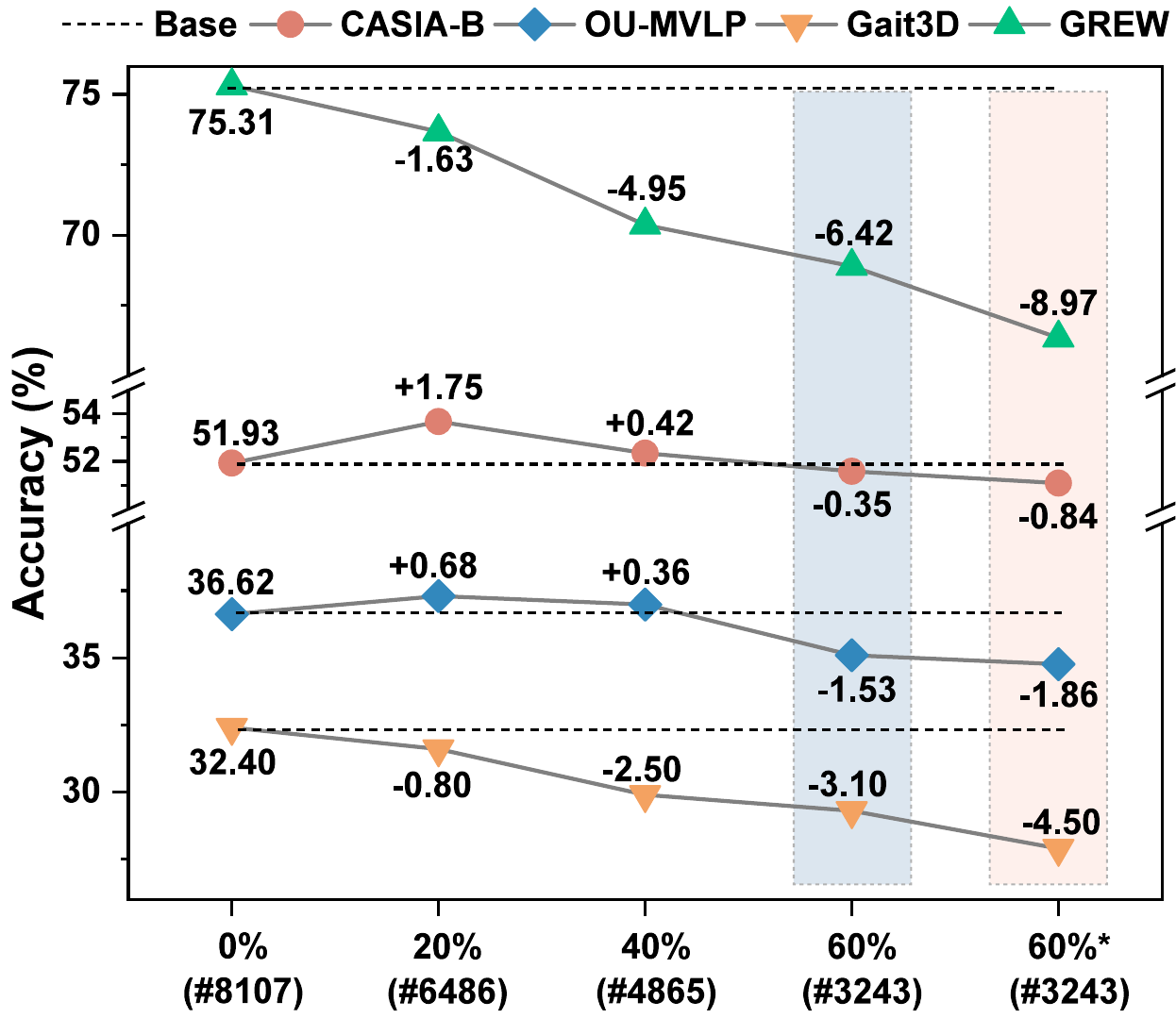}%
    \label{fig:deepgaitv2_grew}%
  }

  \caption{Cross-dataset validation results. 
  The first row represents performance using GaitBase~\cite{fan2023opengait}, while the second row represents performance using DeepGaitV2~\cite{fan2023exploring}. 
  The horizontal axis shows the proportion of low-quality data removed and the number of remaining gait sequences. "-" indicates performance drop, and "+" indicates improvement. "$60\%^*$" means that 60\% of samples are randomly dropped.}
  
  \label{fig:results}
\end{figure*}

Figure~\ref{fig:results} presents the results of the GaitBase~\cite{fan2023opengait} and DeepGaitV2~\cite{fan2023exploring} models trained on different subsets obtained by removing varying proportions of low-quality samples. The impact of dataset distillation is evaluated in both self-domain and cross-domain settings, revealing how data quality influences model performance.

\textbf{Indoor Datasets Results.} For CASIA-B~\cite{CASIA-B}, removing up to 20\% of redundant samples consistently improves both self-domain and cross-domain performance for GaitBase~\cite{fan2023opengait}, with a slight self-domain gain (+0.08\%) and a more noticeable boost when transferring to GREW~\cite{GREW} (+0.45\%). DeepGaitV2~\cite{fan2023exploring} follows a similar pattern, reaching a marginal in-domain improvement (+0.05\%) and better cross-domain accuracy on GREW (+0.30\%). These results indicate that CASIA-B, being relatively clean and balanced, still contains redundant or low-quality sequences that can hinder generalization, and selective removal encourages the model to focus on more informative gait cues. When the removal ratio exceeds 20\%, self-domain accuracy begins to decline for both models, showing that excessive pruning discards useful variations. Interestingly, cross-domain accuracy continues to improve in this regime, implying that overfitting to dataset-specific traits is suppressed. A similar trend emerges in OU-MVLP~\cite{OUMVLP}, where 20\% data reduction preserves strong self-domain performance while significantly boosting cross-domain generalization. GaitBase gains +2.80\% on Gait3D~\cite{Gait3D}, while DeepGaitV2 gains +1.80\%. This demonstrates that OU-MVLP, despite its large scale, also contains sequences that are redundant for within-domain recognition but detrimental for transfer. Notably, random removal yields only modest improvements, whereas targeted distillation achieves more consistent and larger gains, underscoring the effectiveness of quality-based filtering.

\textbf{Outdoor Datasets Results.} For Gait3D~\cite{Gait3D}, eliminating 20\% of noisy samples leads to minor changes in self-domain performance (-0.40\% for GaitBase~\cite{fan2023opengait}, +0.50\% for DeepGaitV2~\cite{fan2023exploring}) but provides measurable cross-domain benefits. DeepGaitV2 achieves +0.83\% improvement when transferring to CASIA-B and +0.56\% to OU-MVLP, which aligns with GaitBase’s trend but at a higher performance level, confirming that distillation reduces the negative impact of cluttered or low-quality samples typical in unconstrained surveillance data. However, when the pruning ratio exceeds 40\%, both models suffer sharp declines in self-domain accuracy (\eg, -6.80\% for GaitBase and -4.80\% for DeepGaitV2), illustrating that outdoor datasets rely heavily on large-scale diversity to capture complex gait variations. GREW~\cite{GREW} shows a parallel phenomenon: 20\% removal causes only slight drops in self-domain results but improves transferability. DeepGaitV2 achieves a cross-domain gain of +1.75\% on CASIA-B compared to +0.82\% with GaitBase, highlighting the former’s stronger generalization ability. Across both datasets, random removal consistently underperforms targeted pruning, further validating that identifying and discarding noisy sequences is crucial to enhancing robustness under domain shift.

\textbf{Mixed Datasets Results.} To further evaluate the generalizability of dataset distillation, we train both GaitBase~\cite{fan2023opengait} and DeepGaitV2~\cite{fan2023exploring} on distilled subsets from multiple datasets (20\% removal ratio), as summarized in Table~\ref{main_results:combined}. Compared to training on the full data, distilled subsets achieve comparable or even better self-domain performance, while consistently enhancing cross-domain accuracy. For instance, GaitBase trained on distilled CASIA-B, Gait3D, and GREW improves cross-domain accuracy on OU-MVLP from 43.22\% to 44.58\% (+1.36\%), alongside a small self-domain gain on Gait3D (63.60\% → 64.00\%). DeepGaitV2 shows even larger benefits: when trained on distilled subsets, it yields +0.70\% improvement on OU-MVLP (48.60\% → 49.30\%) and +0.40\% on Gait3D (76.80\% → 77.20\%). Similarly, with CASIA-B, OU-MVLP, and GREW, distillation improves transfer to Gait3D for both models, from 29.70\% to 30.30\% (GaitBase) and 35.10\% to 35.60\% (DeepGaitV2). These findings demonstrate that distillation remains effective in multi-source training scenarios: pruning redundant or low-quality data reduces noise accumulation from heterogeneous sources while retaining critical diversity. As a result, distilled mixtures produce models that are both more efficient to train and more robust to cross-domain evaluation.

\textbf{Generality across paradigms.}
To further validate that dataset distillation is not confined to our own framework, we extend the experiments to representative methods from two different paradigms. This setting allows us to test whether the benefits of data quality–driven pruning generalize beyond silhouette-based architectures. Specifically, we evaluate (i) the pose-based GPGait~\cite{fu2023gpgait}, which encodes human skeleton sequences for gait recognition, trained on individual datasets (CASIA-B~\cite{CASIA-B}, OUMVLP~\cite{OUMVLP}, Gait3D~\cite{Gait3D}) as well as their combination (CA+OU+G3D, denoted as Mixed), and tested on GREW~\cite{GREW}; and (ii) the RGB-based BigGait~\cite{ye2024biggait}, which leverages full-frame appearance cues, trained on CCPG~\cite{CCPG} and evaluated under cross-dataset transfer to CASIA-B~\cite{CASIA-B} and SUSTech1K~\cite{SUSTech1K}.

For each paradigm, we directly compare three variants: training with the full dataset, with our distilled subsets (removing 20\% of low-quality samples), and with randomly reduced subsets of the same ratio. The results, summarized in Tables~\ref{tab:gpgait_distill} and~\ref{tab:biggait_results}, consistently demonstrate the effectiveness of targeted distillation. In the case of GPGait, distilled subsets provide +0.5\%–1.2\% improvements on GREW, even though the baseline accuracies are relatively low in absolute terms. This shows that pose-based methods, which are highly sensitive to noisy or incomplete skeleton annotations, also benefit from filtering out problematic sequences. For BigGait, distillation improves cross-domain transfer to CASIA-B and SUSTech1K by +1.1\% and +0.7\%, respectively, while random removal leads to performance degradation. Since RGB-based approaches are strongly influenced by background clutter and illumination, removing low-quality samples reduces the risk of overfitting to spurious correlations, thereby yielding more generalizable features.

These observations highlight an important conclusion: the benefits of dataset distillation are not tied to a particular backbone design, input modality, or representation paradigm. Whether operating on silhouettes, poses, or RGB frames, carefully excluding low-quality or redundant samples consistently strengthens cross-domain recognition. This universality suggests that dataset distillation can be regarded as a broadly applicable principle for improving gait recognition systems, serving as a lightweight yet powerful strategy to enhance robustness without altering the model architecture. In addition to accuracy improvements, dataset distillation also reduces training cost. For the most expensive OUMVLP+Gait3D+GREW setting, training time decreases from 33h (full data) to 28h (distilled), while maintaining or even improving accuracy.

\begin{table}[ht]
\centering
\small
\caption{Pose-based GPGait~\cite{fu2023gpgait} trained on different source datasets and evaluated on GREW~\cite{GREW}. 
``*'' denotes distilled subsets. Values in parentheses indicate improvements over full-data training.}
\label{tab:gpgait_distill}
\begin{tabular}{c|cccc}
\toprule
Train Set & CA* & OU* & G3D* & Mixed* \\
\midrule
GREW & 10.7 (+0.7) & 12.3 (+1.2) & 11.6 (+0.6) & 15.4 (+0.5) \\
\bottomrule
\end{tabular}
\end{table}

\begin{table}[ht]
\centering
\small
\caption{Cross-domain evaluation of RGB-based BigGait~\cite{ye2024biggait} trained on CCPG~\cite{CCPG}. 
``Distill-20\%'' denotes our dataset distillation (removing 20\% samples), 
while ``Rand-20\%'' randomly removes 20\%.}
\label{tab:biggait_results}
\begin{tabular}{c|ccc}
\toprule
Test Set & CCPG-Full & Distill-20\% & Rand-20\% \\
\midrule
CASIA-B~\cite{CASIA-B}    & 65.1 & \textbf{66.1} (+1.1) & 64.5 (-0.6) \\
SUSTech1K~\cite{SUSTech1K}  & 64.7 & \textbf{65.4} (+0.7) & 64.3 (-0.4) \\
\bottomrule
\end{tabular}
\end{table}

\textbf{Comparison with Self-supervised Baselines}
To further contextualize cross-domain performance, we compare our method with the self-supervised baseline GaitSSB~\cite{fan2023learning}. 
As shown in Table~\ref{tab:gaitssb_comparison}, our approach consistently outperforms GaitSSB~\cite{fan2023learning} across all datasets, 
demonstrating the effectiveness of dataset distillation even against strong self-supervised methods.

% \\textbf{Cross-domain baseline (self-supervised).}
% We further compare against the self-supervised method GaitSSB to contextualize cross-domain performance under mixed-dataset training.
% Table~\ref{tab:gaitssb_comparison} shows consistent improvements over GaitSSB across all datasets.

\begin{table}[ht]
\centering
\small
\caption{Comparison with GaitSSB~\cite{fan2023learning} under cross-domain settings (Backbone: GaitBase~\cite{fan2023opengait}).}
\label{tab:gaitssb_comparison}
\begin{tabular}{c|cccc}
\toprule
Model  & CASIA-B & OUMVLP & Gai3D & GREW \\
\midrule
GaitSSB & 62.5 & 37.2 & 16.6 & 24.7 \\
\textbf{Ours} & \textbf{63.7} & \textbf{44.6} & \textbf{30.3} & \textbf{32.9} \\
\bottomrule
\end{tabular}
\vspace{-2mm}
\end{table}

\subsection{Ablation Study}
\label{sec:ablation}

\begin{table}[t]
    \centering
    \caption{Ablation study on the effect of Domain-Specific Batch Normalization (DSBN) and Separate Triplet Loss (Se\_Tri) for GaitBase~\cite{fan2023opengait} and DeepGaitV2~\cite{fan2023exploring}. Best and suboptimal results are marked in \textbf{bold} and \underline{underlined}, respectively.}
    \label{ablation_BN_tri_combined}

    \begin{tabular}{cc|c|c|c|c}
        \toprule
        DSBN & Se\_Tri & CA~\cite{CASIA-B} & OU~\cite{OUMVLP} & G3D~\cite{Gait3D} & GREW~\cite{GREW} \\
        \midrule
        \multicolumn{6}{c}{GaitBase~\cite{fan2023opengait}} \\
        \midrule
        \ding{55} & \ding{55} & \underline{90.90} & 84.08 & 62.40 & \underline{32.68} \\
        \checkmark & \ding{55} & 90.33 & \textbf{87.70} & \underline{63.60} & 29.92 \\
        \checkmark & \checkmark & 90.57 & \underline{87.55} & \textbf{64.20} & 31.18 \\
        \ding{55} & \checkmark & \textbf{92.20} & 85.83 & 63.00 & \textbf{33.56} \\
        \midrule
        \multicolumn{6}{c}{DeepGaitV2~\cite{fan2023exploring}} \\
        \midrule
        \ding{55} & \ding{55} & 91.45 & 87.96 & 76.20 & 36.70 \\
        \checkmark & \ding{55} & 91.80 & \textbf{89.20} & 77.50 & 35.40 \\
        \checkmark & \checkmark & \underline{93.12} & \underline{89.05} & \textbf{78.00} & \underline{35.90} \\
        \ding{55} & \checkmark & \textbf{93.63} & 88.35 & \underline{77.60} & \textbf{37.20} \\
        \bottomrule
    \end{tabular}
\end{table}

To further analyze the contribution of key components, we conduct ablation experiments on Domain-Specific Batch Normalization (DSBN)~\cite{chang2019domain} and Separate Triplet Loss (Se\_Tri). The results are summarized in Table~\ref{ablation_BN_tri_combined}.

\textbf{Effect of DSBN.}
Incorporating DSBN generally enhances recognition within the training domain but can hinder cross-domain transfer. For GaitBase~\cite{fan2023opengait}, enabling DSBN substantially improves OU-MVLP~\cite{OUMVLP} accuracy (84.08\% → 87.70\%) and moderately improves Gait3D~\cite{Gait3D} (62.40\% → 63.60\%), indicating that normalizing each dataset with its own statistics helps capture dataset-specific patterns more effectively. Similarly, DeepGaitV2~\cite{fan2023exploring} shows consistent gains on OU-MVLP (87.96\% → 89.20\%) and Gait3D (76.20\% → 77.50\%). However, both models experience notable drops on GREW~\cite{GREW} when DSBN is applied (GaitBase: 32.68\% → 29.92\%; DeepGaitV2: 36.70\% → 35.40\%). This trade-off suggests that while DSBN improves fitting within individual domains, it over-specializes the learned representations, reducing their ability to generalize across unseen distributions.

\textbf{Effect of Separate Triplet Loss (Se\_Tri).}
In contrast, introducing Se\_Tri consistently enhances both self-domain and cross-domain performance. For GaitBase, Se\_Tri increases CASIA-B~\cite{CASIA-B} accuracy from 90.90\% to 92.20\% and improves cross-domain transfer to GREW (32.68\% → 33.56\%). For DeepGaitV2, the improvements are even more pronounced, with CASIA-B rising from 91.45\% to 93.63\% and GREW from 36.70\% to 37.20\%. These results confirm that optimizing triplet losses separately for each domain alleviates gradient conflicts, allowing the model to learn more discriminative, identity-focused features without being dominated by dataset-specific biases. Unlike DSBN, Se\_Tri does not require architectural modifications, making it a lightweight yet effective mechanism for balancing multi-domain optimization.

\textbf{Combined Effect of DSBN and Se\_Tri.}
When DSBN and Se\_Tri are applied together, the models exhibit a mixed pattern. For GaitBase, the combination slightly improves self-domain recognition (e.g., Gait3D: 63.60\% → 64.20\%) but reduces cross-domain robustness compared to using Se\_Tri alone (GREW: 33.56\% → 31.18\%). DeepGaitV2 follows the same trend: the combined setting achieves the highest Gait3D accuracy (78.00\%), indicating that DSBN reinforces dataset-specific fitting, but GREW accuracy drops relative to Se\_Tri alone (37.20\% → 35.90\%). These results suggest that DSBN partially counteracts the generalization benefits of Se\_Tri, as domain-specific normalization introduces dataset fragmentation that limits feature sharing across domains.

\textbf{Summary.}
Overall, the ablation highlights complementary roles: DSBN improves dataset-specific adaptation, while Se\_Tri promotes generalization by decoupling optimization. The best cross-domain performance is achieved by employing Se\_Tri alone, indicating that encouraging discriminative identity learning is more beneficial for robustness than heavily normalizing domain-specific distributions.

%% file: sec/5_conclusion.tex
\section{Conclusion}
\label{sec:conclusion}
This work studies the cross-domain generalization problem in gait recognition under heterogeneous data conditions. We show that improving generalization requires more than simply enlarging training data: supervision inconsistency and sample reliability imbalance remain key obstacles. To address these issues, we develop a unified training framework that enhances supervision reliability and preserves identity-discriminative structure by selectively emphasizing informative samples and employing identity-consistent metric learning. Comprehensive experiments across multiple benchmarks and architectures demonstrate that the proposed approach consistently improves cross-domain performance while maintaining competitive in-domain accuracy. Our analysis further provides practical observations on normalization behaviors in mixed-domain training, offering useful guidance for robust gait representation learning. Future work may extend this direction by exploring adaptive curriculum mechanisms and integrating temporal or contextual priors to further enhance generalizability in complex real-world scenarios.

%% file: ref.bib
@inproceedings{wang2023dygait,
  title={DyGait: Exploiting dynamic representations for high-performance gait recognition},
  author={Wang, Ming and Guo, Xianda and Lin, Beibei and Yang, Tian and Zhu, Zheng and Li, Lincheng and Zhang, Shunli and Yu, Xin},
  booktitle={Proceedings of the IEEE/CVF International Conference on Computer Vision},
  pages={13424--13433},
  year={2023}
}

@inproceedings{GREW,
  title={Gait recognition in the wild: A benchmark},
  author={Zhu, Zheng and Guo, Xianda and Yang, Tian and Huang, Junjie and Deng, Jiankang and Huang, Guan and Du, Dalong and Lu, Jiwen and Zhou, Jie},
  booktitle={Proceedings of the IEEE/CVF international conference on computer vision},
  pages={14789--14799},
  year={2021}
}

@article{guo2025gait,
  title={Gait recognition in the wild: A large-scale benchmark and NAS-based baseline},
  author={Guo, Xianda and Zhu, Zheng and Yang, Tian and Lin, Beibei and Huang, Junjie and Deng, Jiankang and Huang, Guan and Zhou, Jie and Lu, Jiwen},
  journal={IEEE Transactions on Pattern Analysis and Machine Intelligence},
  year={2025},
  publisher={IEEE}
}

@article{wang2025gaitc,
  title={GaitC3I: Robust Cross-Covariate Gait Recognition via Causal Intervention},
  author={Wang, Jilong and Hou, Saihui and Guo, Xianda and Huang, Yan and Huang, Yongzhen and Zhang, Tianzhu and Wang, Liang},
  journal={IEEE Transactions on Circuits and Systems for Video Technology},
  year={2025},
  publisher={IEEE}
}

@inproceedings{wang2022gaitstrip,
  title={Gaitstrip: Gait recognition via effective strip-based feature representations and multi-level framework},
  author={Wang, Ming and Lin, Beibei and Guo, Xianda and Li, Lincheng and Zhu, Zheng and Sun, Jiande and Zhang, Shunli and Liu, Yu and Yu, Xin},
  booktitle={Proceedings of the Asian conference on computer vision},
  pages={536--551},
  year={2022}
}

@article{dou2024clash,
  title={CLASH: Complementary Learning with Neural Architecture Search for Gait Recognition},
  author={Dou, Huanzhang and Zhang, Pengyi and Zhao, Yuhan and Jin, Lu and Li, Xi},
  journal={IEEE Transactions on Image Processing},
  year={2024},
  publisher={IEEE}
}

@article{shen2022comprehensive,
  title={A comprehensive survey on deep gait recognition: algorithms, datasets and challenges},
  author={Shen, Chuanfu and Yu, Shiqi and Wang, Jilong and Huang, George Q and Wang, Liang},
  journal={arXiv preprint arXiv:2206.13732},
  year={2022}
}

@article{sepas2022deep,
  title={Deep gait recognition: A survey},
  author={Sepas-Moghaddam, Alireza and Etemad, Ali},
  journal={IEEE transactions on pattern analysis and machine intelligence},
  volume={45},
  number={1},
  pages={264--284},
  year={2022},
  publisher={IEEE}
}

@article{filipi2022gait,
  title={Gait recognition based on deep learning: a survey},
  author={Filipi Gon{\c{c}}alves dos Santos, Claudio and Oliveira, Diego de Souza and A. Passos, Leandro and Gon{\c{c}}alves Pires, Rafael and Felipe Silva Santos, Daniel and Pascotti Valem, Lucas and P. Moreira, Thierry and Cleison S. Santana, Marcos and Roder, Mateus and Paulo Papa, Jo and others},
  journal={ACM Computing Surveys (CSUR)},
  volume={55},
  number={2},
  pages={1--34},
  year={2022},
  publisher={ACM New York, NY}
}

@inproceedings{chao2019gaitset,
  title={Gaitset: Regarding gait as a set for cross-view gait recognition},
  author={Chao, Hanqing and He, Yiwei and Zhang, Junping and Feng, Jianfeng},
  booktitle={Proceedings of the AAAI conference on artificial intelligence},
  volume={33},
  number={01},
  pages={8126--8133},
  year={2019}
}

@inproceedings{lin2021gait,
  title={Gait recognition via effective global-local feature representation and local temporal aggregation},
  author={Lin, Beibei and Zhang, Shunli and Yu, Xin},
  booktitle={Proceedings of the IEEE/CVF international conference on computer vision},
  pages={14648--14656},
  year={2021}
}

@inproceedings{chang2019domain,
  title={Domain-specific batch normalization for unsupervised domain adaptation},
  author={Chang, Woong-Gi and You, Tackgeun and Seo, Seonguk and Kwak, Suha and Han, Bohyung},
  booktitle={Proceedings of the IEEE/CVF conference on Computer Vision and Pattern Recognition},
  pages={7354--7362},
  year={2019}
}

@inproceedings{fan2023opengait,
  title={Opengait: Revisiting gait recognition towards better practicality},
  author={Fan, Chao and Liang, Junhao and Shen, Chuanfu and Hou, Saihui and Huang, Yongzhen and Yu, Shiqi},
  booktitle={Proceedings of the IEEE/CVF conference on computer vision and pattern recognition},
  pages={9707--9716},
  year={2023}
}

@inproceedings{fan2020gaitpart,
  title={Gaitpart: Temporal part-based model for gait recognition},
  author={Fan, Chao and Peng, Yunjie and Cao, Chunshui and Liu, Xu and Hou, Saihui and Chi, Jiannan and Huang, Yongzhen and Li, Qing and He, Zhiqiang},
  booktitle={Proceedings of the IEEE/CVF conference on computer vision and pattern recognition},
  pages={14225--14233},
  year={2020}
}

@article{ranftl2020towards,
  title={Towards robust monocular depth estimation: Mixing datasets for zero-shot cross-dataset transfer},
  author={Ranftl, Ren{\'e} and Lasinger, Katrin and Hafner, David and Schindler, Konrad and Koltun, Vladlen},
  journal={IEEE transactions on pattern analysis and machine intelligence},
  volume={44},
  number={3},
  pages={1623--1637},
  year={2020},
  publisher={IEEE}
}

@article{shi2024plain,
  title={Plain-Det: A Plain Multi-Dataset Object Detector},
  author={Shi, Cheng and Zhu, Yuchen and Yang, Sibei},
  journal={arXiv preprint arXiv:2407.10083},
  year={2024}
}

@inproceedings{teepe2021gaitgraph,
  title={Gaitgraph: Graph convolutional network for skeleton-based gait recognition},
  author={Teepe, Torben and Khan, Ali and Gilg, Johannes and Herzog, Fabian and H{\"o}rmann, Stefan and Rigoll, Gerhard},
  booktitle={2021 IEEE international conference on image processing (ICIP)},
  pages={2314--2318},
  year={2021},
  organization={IEEE}
}

@inproceedings{fan2024skeletongait,
  title={SkeletonGait: Gait Recognition Using Skeleton Maps},
  author={Fan, Chao and Ma, Jingzhe and Jin, Dongyang and Shen, Chuanfu and Yu, Shiqi},
  booktitle={Proceedings of the AAAI Conference on Artificial Intelligence},
  volume={38},
  number={2},
  pages={1662--1669},
  year={2024}
}

@inproceedings{fu2023gpgait,
  title={Gpgait: Generalized pose-based gait recognition},
  author={Fu, Yang and Meng, Shibei and Hou, Saihui and Hu, Xuecai and Huang, Yongzhen},
  booktitle={Proceedings of the IEEE/CVF International Conference on Computer Vision},
  pages={19595--19604},
  year={2023}
}

@inproceedings{ma2023fine,
  title={Fine-grained unsupervised domain adaptation for gait recognition},
  author={Ma, Kang and Fu, Ying and Zheng, Dezhi and Peng, Yunjie and Cao, Chunshui and Huang, Yongzhen},
  booktitle={Proceedings of the IEEE/CVF International Conference on Computer Vision},
  pages={11313--11322},
  year={2023}
}

@article{fan2023learning,
  title={Learning gait representation from massive unlabelled walking videos: A benchmark},
  author={Fan, Chao and Hou, Saihui and Wang, Jilong and Huang, Yongzhen and Yu, Shiqi},
  journal={IEEE Transactions on Pattern Analysis and Machine Intelligence},
  year={2023},
  publisher={IEEE}
}

@inproceedings{ye2024biggait,
  title={BigGait: Learning Gait Representation You Want by Large Vision Models},
  author={Ye, Dingqiang and Fan, Chao and Ma, Jingzhe and Liu, Xiaoming and Yu, Shiqi},
  booktitle={Proceedings of the IEEE/CVF Conference on Computer Vision and Pattern Recognition},
  pages={200--210},
  year={2024}
}

@article{wang2018dataset,
  title={Dataset distillation},
  author={Wang, Tongzhou and Zhu, Jun-Yan and Torralba, Antonio and Efros, Alexei A},
  journal={arXiv preprint arXiv:1811.10959},
  year={2018}
}

@article{liu2022dataset,
  title={Dataset distillation via factorization},
  author={Liu, Songhua and Wang, Kai and Yang, Xingyi and Ye, Jingwen and Wang, Xinchao},
  journal={Advances in neural information processing systems},
  volume={35},
  pages={1100--1113},
  year={2022}
}

@inproceedings{sun2024diversity,
  title={On the diversity and realism of distilled dataset: An efficient dataset distillation paradigm},
  author={Sun, Peng and Shi, Bei and Yu, Daiwei and Lin, Tao},
  booktitle={Proceedings of the IEEE/CVF Conference on Computer Vision and Pattern Recognition},
  pages={9390--9399},
  year={2024}
}

@inproceedings{sucholutsky2021soft,
  title={Soft-label dataset distillation and text dataset distillation},
  author={Sucholutsky, Ilia and Schonlau, Matthias},
  booktitle={2021 International Joint Conference on Neural Networks (IJCNN)},
  pages={1--8},
  year={2021},
  organization={IEEE}
}

@article{fan2023exploring,
  title={Exploring deep models for practical gait recognition},
  author={Fan, Chao and Hou, Saihui and Huang, Yongzhen and Yu, Shiqi},
  journal={arXiv preprint arXiv:2303.03301},
  year={2023}
}

@inproceedings{BNNecks,
  title={Bag of tricks and a strong baseline for deep person re-identification},
  author={Luo, Hao and Gu, Youzhi and Liao, Xingyu and Lai, Shenqi and Jiang, Wei},
  booktitle={Proceedings of the IEEE/CVF conference on computer vision and pattern recognition workshops},
  pages={0--0},
  year={2019}
}

@inproceedings{HPP,
  title={Horizontal pyramid matching for person re-identification},
  author={Fu, Yang and Wei, Yunchao and Zhou, Yuqian and Shi, Honghui and Huang, Gao and Wang, Xinchao and Yao, Zhiqiang and Huang, Thomas},
  booktitle={Proceedings of the AAAI conference on artificial intelligence},
  volume={33},
  number={01},
  pages={8295--8302},
  year={2019}
}

@inproceedings{CASIA-B,
  title={A framework for evaluating the effect of view angle, clothing and carrying condition on gait recognition},
  author={Yu, Shiqi and Tan, Daoliang and Tan, Tieniu},
  booktitle={18th international conference on pattern recognition (ICPR'06)},
  volume={4},
  pages={441--444},
  year={2006},
  organization={IEEE}
}

@article{OUMVLP,
  title={Multi-view large population gait dataset and its performance evaluation for cross-view gait recognition},
  author={Takemura, Noriko and Makihara, Yasushi and Muramatsu, Daigo and Echigo, Tomio and Yagi, Yasushi},
  journal={IPSJ transactions on Computer Vision and Applications},
  volume={10},
  pages={1--14},
  year={2018},
  publisher={Springer}
}

@inproceedings{Gait3D,
  title={Gait recognition in the wild with dense 3d representations and a benchmark},
  author={Zheng, Jinkai and Liu, Xinchen and Liu, Wu and He, Lingxiao and Yan, Chenggang and Mei, Tao},
  booktitle={Proceedings of the IEEE/CVF conference on computer vision and pattern recognition},
  pages={20228--20237},
  year={2022}
}

@inproceedings{SUSTech1K,
  title={Lidargait: Benchmarking 3d gait recognition with point clouds},
  author={Shen, Chuanfu and Fan, Chao and Wu, Wei and Wang, Rui and Huang, George Q and Yu, Shiqi},
  booktitle={Proceedings of the IEEE/CVF Conference on Computer Vision and Pattern Recognition},
  pages={1054--1063},
  year={2023}
}

@inproceedings{ma2024prompt,
  title={Learning Visual Prompt for Gait Recognition},
  author={Ma, Kang and Fu, Ying and Cao, Chunshui and Hou, Saihui and Huang, Yongzhen and Zheng, Dezhi},
  booktitle={Proceedings of the IEEE/CVF Conference on Computer Vision and Pattern Recognition},
  pages={593--603},
  year={2024}
}

@inproceedings{radford2021learning,
  title={Learning transferable visual models from natural language supervision},
  author={Radford, Alec and Kim, Jong Wook and Hallacy, Chris and Ramesh, Aditya and Goh, Gabriel and Agarwal, Sandhini and Sastry, Girish and Askell, Amanda and Mishkin, Pamela and Clark, Jack and others},
  booktitle={International conference on machine learning},
  pages={8748--8763},
  year={2021},
  organization={PmLR}
}

@inproceedings{jin2025exploring,
  title={Exploring More from Multiple Gait Modalities for Human Identification},
  author={Jin, Dongyang and Fan, Chao and Chen, Weihua and Yu, Shiqi},
  booktitle={Proceedings of the AAAI Conference on Artificial Intelligence},
  volume={39},
  number={4},
  pages={4120--4128},
  year={2025}
}

@article{huang2024gaitdan,
  title={Gaitdan: Cross-view gait recognition via adversarial domain adaptation},
  author={Huang, Tianhuan and Ben, Xianye and Gong, Chen and Xu, Wenzheng and Wu, Qiang and Zhou, Hongchao},
  journal={IEEE Transactions on Circuits and Systems for Video Technology},
  year={2024},
  publisher={IEEE}
}

@inproceedings{zheng2021trand,
  title={Trand: Transferable neighborhood discovery for unsupervised cross-domain gait recognition},
  author={Zheng, Jinkai and Liu, Xinchen and Yan, Chenggang and Zhang, Jiyong and Liu, Wu and Zhang, Xiaoping and Mei, Tao},
  booktitle={2021 IEEE International Symposium on Circuits and Systems (ISCAS)},
  pages={1--5},
  year={2021},
  organization={IEEE}
}

@inproceedings{jaiswal2024domain,
  title={Domain-Specific Adaptation for Enhanced Gait Recognition in Practical Scenarios},
  author={Jaiswal, Nitish and Huan, Vi Duc and Limanta, Felix and Shinoda, Koichi and Wakasa, Masahiro},
  booktitle={Proceedings of the 2024 6th International Conference on Image, Video and Signal Processing},
  pages={8--15},
  year={2024}
}

@article{tong2019gait,
  title={Gait recognition with cross-domain transfer networks},
  author={Tong, Suibing and Fu, Yuzhuo and Ling, Hefei},
  journal={Journal of Systems Architecture},
  volume={93},
  pages={40--47},
  year={2019},
  publisher={Elsevier}
}

@article{schlett2024double,
  title={Double Trouble? Impact and Detection of Duplicates in Face Image Datasets},
  author={Schlett, Torsten and Rathgeb, Christian and Tapia, Juan and Busch, Christoph},
  journal={arXiv preprint arXiv:2401.14088},
  year={2024}
}

@inproceedings{yao2023large,
  title={Large-scale training data search for object re-identification},
  author={Yao, Yue and Gedeon, Tom and Zheng, Liang},
  booktitle={Proceedings of the IEEE/CVF Conference on Computer Vision and Pattern Recognition},
  pages={15568--15578},
  year={2023}
}

@inproceedings{gowda2021smart,
  title={Smart frame selection for action recognition},
  author={Gowda, Shreyank N and Rohrbach, Marcus and Sevilla-Lara, Laura},
  booktitle={Proceedings of the AAAI Conference on Artificial Intelligence},
  volume={35},
  number={2},
  pages={1451--1459},
  year={2021}
}

@InProceedings{CCPG,
    author    = {Li, Weijia and Hou, Saihui and Zhang, Chunjie and Cao, Chunshui and Liu, Xu and Huang, Yongzhen and Zhao, Yao},
    title     = {An In-Depth Exploration of Person Re-Identification and Gait Recognition in Cloth-Changing Conditions},
    booktitle = {Proceedings of the IEEE/CVF Conference on Computer Vision and Pattern Recognition (CVPR)},
    month     = {June},
    year      = {2023},
    pages     = {13824-13833}
}

@article{wang2023gait,
  title={Gait recognition with multi-level skeleton-guided refinement},
  author={Wang, Runsheng and Shi, Yuxuan and Ling, Hefei and Li, Zongyi and Zhao, Chengxin and Wei, Bohao and Li, He and Li, Ping},
  journal={IEEE Transactions on Multimedia},
  volume={26},
  pages={4515--4526},
  year={2023},
  publisher={IEEE}
}

@article{yao2022improving,
  title={Improving disentangled representation learning for gait recognition using group supervision},
  author={Yao, Lingxiang and Kusakunniran, Worapan and Zhang, Peng and Wu, Qiang and Zhang, Jian},
  journal={IEEE Transactions on Multimedia},
  volume={25},
  pages={4187--4198},
  year={2022},
  publisher={IEEE}
}

@article{zhao2021associated,
  title={Associated spatio-temporal capsule network for gait recognition},
  author={Zhao, Aite and Dong, Junyu and Li, Jianbo and Qi, Lin and Zhou, Huiyu},
  journal={IEEE Transactions on Multimedia},
  volume={24},
  pages={846--860},
  year={2022},
  publisher={IEEE}
}

@ARTICLE{10243069,
  author={Li, Aoqi and Hou, Saihui and Cai, Qingyuan and Fu, Yang and Huang, Yongzhen},
  journal={IEEE Transactions on Multimedia}, 
  title={Gait Recognition With Drones: A Benchmark}, 
  year={2024},
  volume={26},
  number={},
  pages={3530-3540},
}

@article{sun2024dualistic,
  title={Dualistic Disentangled Meta-Learning Model for Generalizable Person Re-Identification},
  author={Sun, Jia and Li, Yanfeng and Chen, Luyifu and Chen, Houjin and Wang, Minjun},
  journal={IEEE Transactions on Information Forensics and Security},
  year={2024},
  publisher={IEEE}
}

@inproceedings{deng2019arcface,
  title={Arcface: Additive angular margin loss for deep face recognition},
  author={Deng, Jiankang and Guo, Jia and Xue, Niannan and Zafeiriou, Stefanos},
  booktitle={Proceedings of the IEEE/CVF conference on computer vision and pattern recognition},
  pages={4690--4699},
  year={2019}
}

@article{lin2024human,
  title={Human-centric transformer for domain adaptive action recognition},
  author={Lin, Kun-Yu and Zhou, Jiaming and Zheng, Wei-Shi},
  journal={IEEE Transactions on Pattern Analysis and Machine Intelligence},
  year={2024},
  publisher={IEEE}
}

@article{zhou2022domain,
  title={Domain generalization: A survey},
  author={Zhou, Kaiyang and Liu, Ziwei and Qiao, Yu and Xiang, Tao and Loy, Chen Change},
  journal={IEEE transactions on pattern analysis and machine intelligence},
  volume={45},
  number={4},
  pages={4396--4415},
  year={2022},
  publisher={IEEE}
}

@article{wang2022generalizing,
  title={Generalizing to unseen domains: A survey on domain generalization},
  author={Wang, Jindong and Lan, Cuiling and Liu, Chang and Ouyang, Yidong and Qin, Tao and Lu, Wang and Chen, Yiqiang and Zeng, Wenjun and Yu, Philip S},
  journal={IEEE transactions on knowledge and data engineering},
  volume={35},
  number={8},
  pages={8052--8072},
  year={2022},
  publisher={IEEE}
}

@ARTICLE{huang2025watch,
  author={Huang, Binyuan and Luo, Yongdong and Guo, Xianda and Zheng, Xiawu and Zhu, Zheng and Pan, Jiahui and Zhou, Chengju},
  journal={IEEE Transactions on Multimedia}, 
  title={Watch Where You Move: Region-Aware Dynamic Aggregation and Excitation for Gait Recognition}, 
  year={2025},
  volume={},
  number={},
  pages={1-12},
}
